\def\year{2019}\relax
\documentclass[letterpaper]{article} %DO NOT CHANGE THIS
\usepackage{aaai19}  %Required
\usepackage{times}  %Required
\usepackage{helvet}  %Required
\usepackage{courier}  %Required
\def\year{2019}\relax
\usepackage{aaai19}  %Required
\usepackage{times}  %Required
\usepackage{helvet}  %Required
\usepackage{courier}  %Required
\usepackage{url}  %Required
\usepackage{amsmath, amssymb, epsfig, graphicx, bm}
\usepackage{cases}
\usepackage{float, subfigure}
\usepackage{color}
\usepackage{amsthm}
\usepackage{amsfonts}
\usepackage{algorithmic,algorithm}
\usepackage{graphicx}  %Required
\newtheorem{theorem}{Theorem}
\newtheorem{proposition}{Proposition}

\newtheorem{remark}{Remark}

\newtheorem{assumption}{Assumption}

\newcommand{\R}{\mathcal{R}}
\newcommand{\Rb}{\mathbb{R}}

\newcommand{\x}{\tilde{x}}

\newcommand{\n}{\nabla}
\newcommand{\f}{\frac}
\newcommand{\la}{\lambda}
\newcommand{\E}{\mathbb{E}}
\newcommand{\A}{\alpha}
\newcommand{\La}{\langle}
\newcommand{\Ra}{\rangle}
\newcommand{\D}{\Delta_2}

\definecolor{rev}{rgb}{1, 0.2, 0.2}

\begin{document}
\title{RSA: Byzantine-Robust Stochastic Aggregation Methods for \\ Distributed Learning from Heterogeneous Datasets}
%\author{
%Anonymous
%}
\author{Liping Li$^*$~~~~~~ Wei Xu$^*$~~~~~~ Tianyi Chen$^{\dagger}$~~~~~~ Georgios B. Giannakis$^{\dagger}$~~~~~~ Qing Ling$^{\ddagger}$\\
\\
$^*$Department of Automation, University of Science and Technology of China, Hefei, Anhui, China\\
$^{\dagger}$Digital Technology Center, University of Minnesota, Twin Cities, Minneapolis, Minnesota, USA\\
$^{\ddagger}$School of Data and Computer Science, Sun Yat-Sen University, Guangzhou, Guangdong, China\\
}
\vspace{5em}

\maketitle
\begin{abstract}
In this paper, we propose a class of robust stochastic subgradient methods for distributed learning from heterogeneous datasets at presence of an unknown number of Byzantine workers. The Byzantine workers, during the learning process, may send arbitrary incorrect messages to the master due to data corruptions, communication failures or malicious attacks, and consequently bias the learned model. The key to the proposed methods is a regularization term incorporated with the objective function so as to robustify the learning task and mitigate the negative effects of Byzantine attacks. The resultant subgradient-based algorithms are termed \textit{Byzantine-Robust Stochastic Aggregation} methods, justifying our acronym RSA used henceforth. In contrast to most of the existing algorithms, RSA does not rely on the assumption that the data are independent and identically distributed (i.i.d.) on the workers, and hence fits for a wider class of applications.
%{\color{blue}Theoretically, we prove that the convergence rate of RSA under Byzantine attacks}
{Theoretically, we show that: i) RSA converges to a near-optimal solution with the learning error dependent on the number of Byzantine workers; ii) the convergence rate of RSA under Byzantine attacks is the same as that of the stochastic gradient descent method, which is free of Byzantine attacks.}
Numerically, experiments on real dataset corroborate the competitive performance of RSA and a complexity reduction compared to the state-of-the-art alternatives.
\end{abstract}

\section{Introduction}
The past decade has witnessed the proliferation of smart phones and Internet-of-Things (IoT) devices. They generate a huge amount of data every day, from which one can learn models of cyber-physical systems and make decisions to improve the welfare of human being. Nevertheless, standard machine learning approaches that require centralizing the training data on one machine or in a datacenter may not be suitable for such applications, as data collected from distributed devices and stored at clouds lead to significant privacy risks \cite{sicari2015}. To alleviate user privacy concerns, a new distributed machine learning framework called \emph{federated learning} has been proposed by Google and become popular recently \cite{mcmahan2017blog,smith2017}. Federated learning allows the training data to be kept locally on the owners' devices. Data samples and computation tasks are distributed across multiple workers such as Internet-of-Things (IoT) devices in a smart home, which are programmed to collaboratively learn a model. Parallel implementations of popular machine learning algorithms, such as stochastic gradient descent (SGD), are applied to learning from the distributed data \cite{bottou2010}.

However, federated learning still faces two significant challenges: high communication overhead and serious security risk. While several recent approaches have been developed to tackle the communication bottleneck of distributed learning \cite{li2014,liu2017,smith2017,chen2018lag}, the security issue has not been adequately addressed. In federated learning applications, a number of devices may be highly unreliable or even easily compromised by hackers. We call these devices as Byzantine workers. In this scenario, the learner lacks secure training ability, which makes it vulnerable to failures, not mentioning adversarial attacks \cite{lynch1996}. For example, SGD, the workhorse of large-scale machine learning, is vulnerable to even one Byzantine worker \cite{geomatric}.

In this context, the present paper studies distributed machine learning under a general Byzantine failure model, where the Byzantine workers can arbitrarily modify the messages transmitted from themselves to the master. With such a model, it simply does not have any constraints on the communication failures or attacks. We aim to develop efficient  distributed machine learning methods tailored for this setting with provable performance guarantee.

\subsection{Related work}
Byzantine-robust distributed learning has received increasing attention in recent years. Most of the existing algorithms extend SGD to incorporate the Byzantine-robust setting and assume that the data are independent and identically distributed (i.i.d.) on the workers. Under this assumption, stochastic gradients computed by regular workers are presumably distributed around the true gradient, while those sent from the Byzantine workers to the master could be arbitrary. Thus, the master is able to apply robust estimation techniques to aggregate the stochastic gradients. Typical gradient aggregation rules include geometric median \cite{geomatric},  marginal trimmed mean \cite{trimmed_mean,Phocas}, dimensional median \cite{genneral-BSGD,Ali-threshold}, etc. A more sophisticated algorithm termed as Krum selects a gradient which has minimal summation of Euclidean distances from a given number of nearest gradients \cite{Krum}. Targeting high-dimensional learning, an iterative filtering algorithm is developed in \cite{Su-Secure-high-dimension}, which achieves the optimal error rate in the high-dimensional regime. The main disadvantage of these existing algorithms comes from the i.i.d. assumption, which is arguably not the case in federated learning over heterogeneous computing units. Actually, generalizing these algorithms to the non-i.i.d. setting is not straightforward. In addition, some of these algorithms rely on sophisticated gradient selection subroutines, such as those in Krum and geometric median, which incur high computational complexity.

Other related work in this context includes \cite{escape_saddle_point} that targets escaping saddle points of nonconvex optimization problems under Byzantine attacks, and \cite{DRACO} that leverages a gradient-coding based algorithm for robust learning. However, the approach in \cite{DRACO} needs to relocate the data points, which is not easy to implement in the federated learning paradigm. Leveraging additional data, \cite{Zeno} studies the trustworthy score-based schemes that guarantee efficient learning even when there is only one non-Byzantine worker, but additional data may not always be available in practice. Our algorithms are also related to robust decentralized optimization studied in, e.g., \cite{ben2016robust,xu2018}, which consider optimizing a static or dynamic cost function over a decentralized network with unreliable nodes. In contrast, the focus of this work is Byzantine-robust stochastic optimization.

\subsection{Our contributions}
The contributions of this paper are summarized as follows.

c1) We develop a class of robust stochastic methods abbreviated as RSA for distributed learning over heterogeneous datasets and under Byzantine attacks. RSA has several variants, each tailored for an $\ell_p$-norm regularized robustifying objective function.

c2) Performance is rigorously established for the resultant RSA approaches, in terms of the convergence rate as well as the error caused by the Byzantine attacks.

c3) Extensive numerical tests using the MNIST dataset are conducted to corroborate the effectiveness of RSA in term of both classification accuracy under Byzantine attacks and runtime.

\section{Distributed SGD}
We consider a general distributed system, consisting of a master and $m$ workers, among which $q$ workers are Byzantine (behaving arbitrarily). The goal is to find the optimizer of the following problem:
\begin{align}\label{eq:obj}
\min\limits_{\tilde{x}\in\mathbb{R}^d} \sum_{i=1}^m\mathbb{E}[F(\tilde{x},\xi_i)]+ f_0(\tilde{x}).
\end{align}
Here $\tilde{x}\in\mathbb{R}^d$ is the optimization variable, $f_0(\tilde{x})$ is a regularization term, and $F(\tilde{x},\xi_i)$ is the loss function of worker $i$ with respect to a random variable $\xi_i$. Unlike the previous work which assumes the distributed data across the workers are i.i.d., we consider a more practical situation: $\xi_i\sim\mathcal{D}_i$, where $\mathcal{D}_i$ is the data distribution on worker $i$ and could be different to the distributions on other workers.

In the master-worker architecture, at time $k+1$ of the distributed SGD algorithm, every worker $i$ receives the current model $\tilde{x}^k$ from the master, samples a data point from the distribution $\mathcal{D}_i$ with respect to a random variable $\xi_i^k$, and computes the gradient of the local empirical loss $\nabla F(\tilde{x}^k,\xi_i^k)$. Note that this sampling process can be easily generalized to the mini-batch setting, in which every worker samples multiple i.i.d. data points and computes the averaged gradient of the local empirical losses. The master collects and aggregates the gradients sent by the workers, and updates the model. Its update at time $k+1$ is:
\begin{align}\label{eq:update:x}
\!\!\tilde{x}^{k+1} = \tilde{x}^k-\alpha^{k+1}\!\left(\nabla f_0(\tilde{x}^k)+\sum_{i=1}^m \nabla F(\tilde{x}^k,\xi_i^k)\right)
\end{align}
where $\alpha^{k+1}$ is a diminishing learning rate at time $k+1$. The distributed SGD is outlined in Algorithm \ref{algo0}.

\begin{algorithm}[t]
	\caption{Distributed SGD}\label{algo0}
	\begin{algorithmic}[1]
		\centerline{\textbf{Master}:}
		\STATE Input: $\tilde{x}^0$, $\alpha^k$. At time $k+1$:
		\STATE Broadcast its current iterate $\tilde{x}^k$ to all workers;
		\STATE Receive all gradients $\nabla F(\tilde{x}^k,\xi_i^k)$ sent by workers;
		\STATE Update the iterate via \eqref{eq:update:x}.
	\end{algorithmic}
	\vspace{0.3cm}
	\begin{algorithmic}[1]
		\centerline{\textbf{Worker} $i$:}
		\STATE At time $k+1$:
		\STATE Receive the master's current iterate $\tilde{x}^k$;
		\STATE Compute a local stochastic gradient $\nabla F(\tilde{x}^k,\xi_i^k)$;
		\STATE Send the local stochastic gradient to the server.
	\end{algorithmic}
\end{algorithm}

\noindent\textbf{SGD is vulnerable to Byzantine attacks}.
While SGD has well-documented performance in conventional large-scale machine learning settings, its performance will significantly degrade at the presence of Byzantine workers \cite{geomatric}. Suppose that some of the workers are Byzantine, they can report arbitrary messages or strategically send well-designed messages according to the information sent by other workers so as to bias the learning process. Specifically, if worker $m$ is Byzantine, at time $k+1$, it can choose one of two following attacks:

a1) sending $\nabla F(\tilde{x}^k,\xi_m^k)=\infty$; %and,

a2) sending $\nabla F(\tilde{x}^k,\xi_m^k)=-\sum_{i=1}^{m-1} \nabla F(\tilde{x}^k,\xi_i^k)$.

\begin{figure}[t]
	\begin{center}
		\includegraphics[scale=0.45]{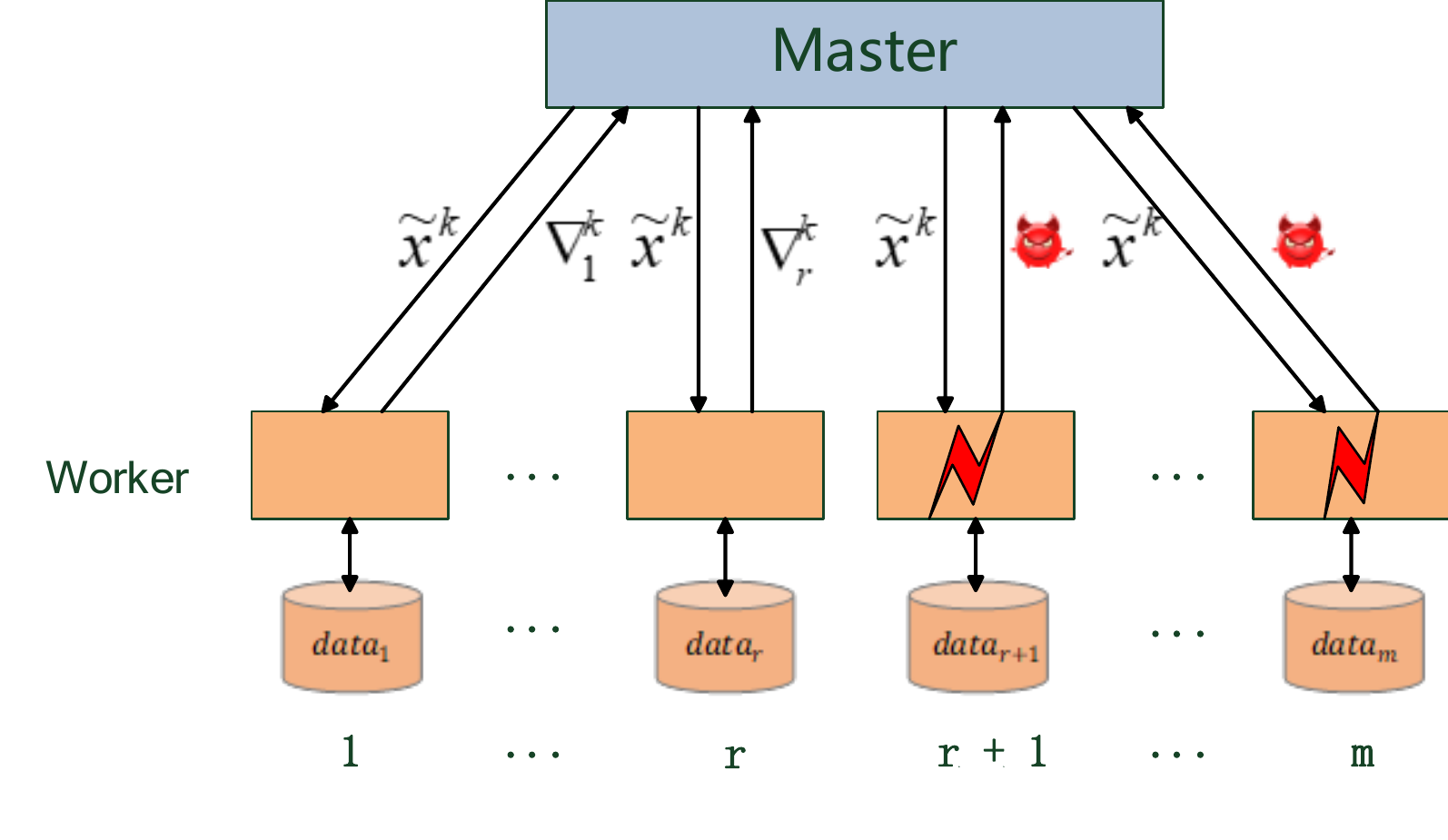}
		\caption{The operation of SGD. There are $m$ workers, $r$ being regular and the rest of $q=m-r$ being Byzantine. The master sends the current iterate to the workers, and the regular workers send back the stochastic gradients. The red devil marks denote the wrong messages that the Byzantine workers send to the master.}\label{eps:SGD_PS}
	\end{center}
	\vspace{-0.5cm}
\end{figure}

\noindent In any case, the aggregated gradient $\sum_{i=1}^m \nabla F(\tilde{x}^k,\xi_i^k)$ used in the SGD update \eqref{eq:update:x} will be either infinite or null, and thus the learned model $\tilde{x}^k$ will either not converge or converge to an incorrect value. The operation of SGD under Byzantine attacks is illustrated in Figure \ref{eps:SGD_PS}.

Instead of using the simple averaging in \eqref{eq:update:x}, robust gradient aggregation rules have been incorporated with SGD in \cite{Krum,geomatric,genneral-BSGD,escape_saddle_point,trimmed_mean,Zeno}. However, in the federated learning setting, these aggregation rules become less effective due to the difficulty of distinguishing the statistical heterogeneity from the Byzantine attacks. In what follows, we develop a counterpart of SGD to address the issue of robust learning from distributed heterogeneous data.

\section{RSA for Robust Distributed Learning}
Observe that due to the presence of the Byzantine workers, it is meaningless to solve \eqref{eq:obj}, which minimizes the summation of all workers' local expected losses without distinguishing regular and Byzantine workers, because the Byzantine workers can always prevent the learner from accessing their local data and finding the optimal solution. Instead, a less ambitious goal is to find a solution that minimizes the summation of the regular workers' local expected cost functions plus the regularization term:
\begin{align}\label{eq:correct}
\tilde{x}^* = \arg\min\limits_{\tilde{x}\in\mathbb{R}^d} \sum_{i\in\mathcal{R}}\mathbb{E}[F(\tilde{x},\xi_i)]
+ f_0(\tilde{x})
\end{align}
Here we denote $\mathcal{B}$ as the set of Byzantine workers and $\mathcal{R}$ as the set of regular workers, with $|\mathcal{B}|=q$ and $|\mathcal{R}|=m-q$. Letting each regular worker $i$ have its local iterate $x_i$ and the master have its local iterate $x_0$, we obtain an equivalent form to \eqref{eq:correct}:
\begin{subequations}\label{eq:correct2}
	\begin{align}
	\min\limits_{x:=[x_i;x_0]}&~\sum_{i\in\mathcal{R}}\mathbb{E}[F(x_i,\xi_i)] + f_0(x_0)\\
	{\rm s. to}&~~~ x_0=x_i, ~ \forall i \in \mathcal{R}
	\end{align}
\end{subequations}
where $x:=[x_i;x_0]\in\mathbb{R}^{(|\mathcal{R}|+1)d}$ is a vector that stacks the regular workers' local variables $x_i$ and the master's variable $x_0$. The formulation \eqref{eq:correct2} is aligned with the concept of consensus optimization in, e.g., \cite{shi2014admm}.

\subsection{$\ell_1$-norm RSA}
Directly solving \eqref{eq:correct} or \eqref{eq:correct2} by iteratively updating $\tilde{x}$ or $x$ is impossible since the identities of Byzantine workers are not available to the master. Therefore, we introduce an $\ell_1$-norm regularized form of \eqref{eq:correct2}:
\begin{equation}\label{eq:tv}
x^*\!:=\!\mathop{\arg\min}\limits_{x:= [x_i;x_0]}\sum_{i\in\mathcal{R}}\Big(\mathbb{E}[F(x_i,\xi_i)]+\lambda||x_i-x_0||_1\Big)+ f_0(x_0)
\end{equation}
where $\lambda$ is a positive constant. The second term in the cost function \eqref{eq:tv} is the $\ell_1$-norm penalty, whose minimization forces every $x_i$ to be close to the master's variable $x_0$. We will show next that how this relaxed form brings the advantage of robust learning under Byzantine attacks.

In the ideal case that the identities of Byzantine workers are revealed, we can apply a stochastic subgradient method to solve \eqref{eq:tv}. The optimization only involves the regular workers and the master. At time $k+1$, the updates of $x_i^{k+1}$ at regular worker $i$ and $x_0^{k+1}$ at the master are given by:
\begin{subequations}\label{eq:diff:x}
	\begin{align}
	& \hspace{-1em} x_i^{k+1} = x_i^k \!-\!\alpha^{k+1}\left(\nabla F(x_i^k,\xi_i^k)\!+\!\lambda {\rm sign}(x_i^k\!-\!x_0^k)\right)\label{eq:diff:x-1}\\
	& \hspace{-1em} x_0^{k+1} = x_0^k \!-\!\alpha^{k+1}\Big(\nabla f_0(x_0^k)\!+\!\lambda\Big(\sum_{i\in\mathcal{R}}{\rm sign}(x_0^k\!-\!x_i^k)\!\Big)\!\Big)\label{eq:diff:x-2}
	\end{align}
\end{subequations}
where ${\rm sign}(\cdot)$ is the element-wise sign function. Given $a\in\mathbb{R}, {\rm sign}(a)$ equals to $1$ when $a > 0$, $-1$ when $a < 0$, and an arbitrary value within $[-1, 1]$ when $a=0$. At time $k+1$, each worker $i$ sends the local iterate $x_i^k$ to the master, instead of sending its local stochastic gradient in distributed SGD. The master aggregates the models sent by the workers to update its own model $x_0^{k+1}$. In this sense, the updates in \eqref{eq:diff:x} are based on model aggregation, different to gradient aggregation in SGD.

Now let us consider how the updates in \eqref{eq:diff:x} behave at presence of Byzantine workers. The update of a regular worker $i$ is the same as \eqref{eq:diff:x-1}, which is:
\begin{align}\label{eq:update:x:i}
x_i^{k+1} = x_i^{k}\!-\!\alpha^{k+1}\left(\nabla F(x_i^k,\xi_i^k)\!+\!\lambda {\rm sign}(x_i^k\!-\!x_0^k)\right).
\end{align}
If worker $i$ is Byzantine, instead of sending the value $x_i^k$ computed from \eqref{eq:diff:x-1} to the master, it sends an arbitrary variable $z_i^k\in\mathbb{R}^d$. The master is unable to distinguish $x_i^k$ sent by a regular worker or $z_i^k$ sent by a Byzantine worker. Therefore, the update of the master at time $k+1$ is no longer \eqref{eq:diff:x-2}, but:
\begin{align}\label{eq:update:x:0}
x_0^{k+1} = x_0^k \!-\!\alpha^{k+1}\Big(&\nabla f_0(x_0^k)+\lambda\Big(\sum_{i\in\mathcal{R}}{\rm sign}(x_0^k-x_i^k)\nonumber\\
&+\sum_{j\in\mathcal{B}}{\rm sign}(x_0^k-z_j^k)\Big)\Big).
\end{align}
We term this algorithm as $\ell_1$-norm RSA (Byzantine-robust stochastic aggregation).

\noindent\textbf{$\ell_1$-norm RSA is robust to Byzantine attacks}. $\ell_1$-norm RSA is robust to Byzantine attacks due to the introduction of the $\ell_1$-norm regularized term to \eqref{eq:tv}. The regularization term allows every $x_i$ to be different from $x_0$, and the bias is controlled by the parameter $\lambda$. This modification robustifies the objective function when any worker is Byzantine and behaves arbitrarily. From the algorithmic perspective, we can observe from the update \eqref{eq:update:x:0} that the impacts of a regular worker and a Byzantine worker on $x_0^{k+1}$ are similar, no matter the how different the values sent by them to the master are. Therefore, only the number of Byzantine workers will influence the RSA update \eqref{eq:update:x:0}, rather than the magnitudes of malicious messages sent by the Byzantine workers. In this sense, $\ell_1$-norm RSA is robust to arbitrary attacks from Byzantine workers. This is in sharp comparison with SGD, which is vulnerable to even a single Byzantine worker.

\subsection{Generalization to $\ell_p$-norm RSA}
In addition to solving $\ell_1$-norm regularized problem \eqref{eq:tv}, we can also solve the following $\ell_p$-norm regularized problem:
\begin{equation}\label{eq:tv-p}
x^*\!:=\!\mathop{\arg\min}\limits_{x:= [x_i;x_0]}\sum_{i\in\mathcal{R}}\Big(\mathbb{E}[F(x_i,\xi_i)]+\lambda||x_i-x_0||_p\Big)+ f_0(x_0)
\end{equation}
where $p \geq 1$. Similar to the case of $\ell_1$-regularized objective in \eqref{eq:tv}, the $\ell_p$ norm penalty helps mitigate the negative influence of the Byzantine workers.

Akin to the $\ell_1$-norm RSA, the $\ell_p$-norm RSA still operates using subgradient recursions. For each regular worker $i$, its local update at time $k+1$ is:
\begin{align}\label{eq:update2:x:i}
\hspace{-1em} x_i^{k+1} = x_i^k\!-\!\alpha^{k+1}\left(\nabla F(x_i^k,\xi_i^k)+\lambda \partial_{x_i} \|x_i^k-x_0^k\|_p\right)
\end{align}
where $\partial_{x_i} \|x_i^k-x_0^k\|_p$ is a subgradient of $\|x_i-x_0^k\|_p$ at $x_i = x_i^k$. Likewise, for the master, its update at time $k+1$ is:
\begin{align}\label{eq:update2:x:0}
x_0^{k+1} = x_0^k\!-\!\alpha^{k+1}\Big(&\nabla f_0(x_0^k)+\lambda\Big(\sum_{i\in\mathcal{R}}\partial_{x_0} \|x_0^k-x_i^k\|_p\nonumber\\
&+\sum_{j\in\mathcal{B}}\partial_{x_0} \|x_0^k-z_j^k\|_p\Big)\Big)
\end{align}
where $\partial_{x_0} \|x_0^k-x_i^k\|_p$ and $\partial_{x_0} \|x_0^k-z_j^k\|_p$ are subgradients of $\|x_0-x_j^k\|_p$ and $\|x_0-z_j^k\|_p$ at $x_0 = x_0^k$, respectively.

To compute the subgradient involved in $\ell_p$-norm RSA, we will rely on the following proposition.
\begin{proposition}\label{prop.1}
	Let $p \geq 1$ and $b$ satisfy $\f{1}{b}+\f{1}{p}=1$. For $x\in\Rb^d$, we have the subdifferential $\partial_x \|x\|_p=\{z\in \Rb^d: \La z,x\Ra=\|x\|_p,\ \|z\|_b\leq 1\}$.
\end{proposition}
Here and thereafter, we slightly abuse the notation by using $\partial$ to denote both subgradient and subdifferential. The proof of Proposition \ref{prop.1} is in the supplementary document.

\begin{figure}[t]
	\begin{center}
		\includegraphics[scale=0.45]{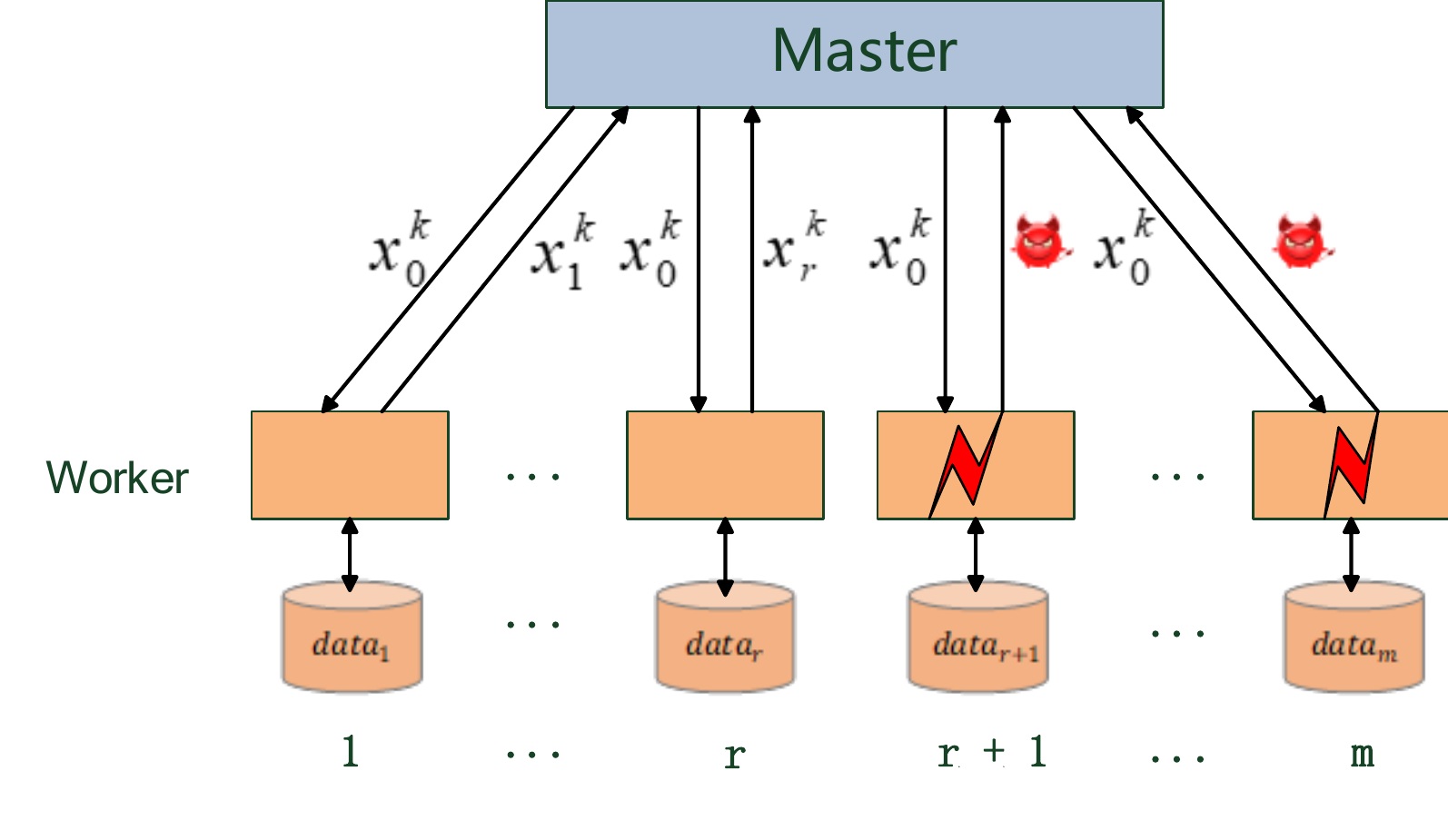}
		\caption{The operation of RSA. There are $m$ workers, $r$ being regular and the rest of $q=m-r$ being Byzantine. The master sends its local variable to the workers, and the regular workers send back their local variables. The red devil marks denote the wrong messages that the Byzantine workers send to the master.}\label{eps:PS2}
	\end{center}
	\vspace{-0.5cm}
\end{figure}

Together with $\ell_1$-norm RSA, $\ell_p$-norm RSA for robust distributed stochastic optimization under Byzantine attacks is summarized in Algorithm \ref{algo1} and illustrated in Figure \ref{eps:PS2}.

\begin{algorithm}[t]
	\caption{RSA for Robust Distributed Learning}\label{algo1}
	\begin{algorithmic}[1]
		\centerline{\textbf{Master:}}
		\STATE Input: $x_0^0$, $\lambda>0$, $\alpha^k$. At time $k+1$:
		\STATE Broadcast its current iterates $x_0^k$ to all workers;
		\STATE Receive all local iterates $x_i^k$ sent by regular workers or faulty values $z_i^k$ sent by Byzantine workers;
		\STATE Update the iterate via \eqref{eq:update:x:0} or \eqref{eq:update2:x:0}.
	\end{algorithmic}
	\vspace{0.3cm}
	\begin{algorithmic}[1]
		\centerline{\textbf{Regular Worker} $i$:}
		\STATE Input: $x_i^0$, $\lambda>0$, $\alpha^k$. At time $k+1$:
		\STATE Send the current local iterate $x_i^k$ to the master;
		\STATE Receive the master's local iterate $x_0^k$;
		\STATE Update the local iterate via \eqref{eq:update:x:i} or \eqref{eq:update2:x:i}.
	\end{algorithmic}
\end{algorithm}

\begin{remark}[Model vs. gradient aggregation]\label{remark 1}
	Most existing Byzantine-robust methods are based on gradient aggregation. Since each worker computes its gradient using the same iterate, these methods do not have the consensus issue \cite{Krum,geomatric,genneral-BSGD,escape_saddle_point,trimmed_mean,Zeno}. However, to enable efficient gradient aggregation, these methods require the data stored in the workers are i.i.d., which is impractical in the federated learning setting. {Under the assumption that data are i.i.d. on all the workers, the stochastic gradients computed by regular workers are also i.i.d. and their expectations are equal. Using this prior knowledge, the master is able to apply robust estimation techniques to aggregate the stochastic gradients collected from both regular and Byzantine workers. When the i.i.d. assumption does not hold, even the stochastic gradients computed by regular workers are different in expectation, such that the gradient aggregation-based methods no longer work. In comparison, the proposed RSA methods utilize model aggregation aiming at finding a consensual model, and do not rely on the i.i.d. assumption. }On the other hand, the existing gradient aggregation methods generally require to design nontrivial subroutines to aggregate gradients, and hence incur relatively high complexities \cite{genneral-BSGD,Krum,Su-Secure-high-dimension}. In contrast, the proposed RSA methods enjoy much lower complexities, which are the same as that of the standard distributed SGD working in the Byzantine-free setting. We shall demonstrate the advantage of RSA in computational time in the numerical tests, in comparison with several state-of-the-art alternatives.
\end{remark}

\section{Convergence Analysis}

This section analyzes the performance of the proposed RSA methods, with proofs given in the supplementary document. We make the following assumptions.
\begin{assumption}\label{ass:conv}
	\textbf{\emph{(Strong convexity)}} \emph{The local cost functions $\mathbb{E}[F(\tilde{x},\xi_i)]$ and the regularization term $f_0(\tilde{x})$ are strongly convex with constants $\mu_i$ and $\mu_0$, respectively}.
\end{assumption}
\begin{assumption}\label{ass:lip}
	\textbf{\emph{(Lipschitz continuous gradients)}} \emph{The local cost functions $\mathbb{E}[F(\tilde{x},\xi_i)]$ and the regularization term $f_0(\tilde{x})$ have Lipschitz continuous gradients with constants $L_i$ and $L_0$, respectively}.
\end{assumption}
\begin{assumption}\label{ass:bound}
	\textbf{\emph{(Bounded variance)}} \emph{For every worker $i$, the data sampling is i.i.d. across time such that $\xi_i^k \sim \mathcal{D}_i$. The variance of $\nabla F(\tilde{x},\xi_i)$ is upper bounded by $\delta_i^2$, namely, $\E[||\nabla \mathbb{E}[F(\tilde{x},\xi_i)]- \nabla F(\tilde{x},\xi_i)||^2] \leq \delta_i^2$}.
\end{assumption}
Note that Assumptions 1-3 are standard for performance analysis of stochastic gradient-based methods \cite{nemirovski2009}, and they are satisfied in a wide range of machine learning problems such as $\ell_2$-regularized least squares and logistic regression.

We start with investigating the $\ell_p$-norm regularized problem \eqref{eq:tv-p}, showing the condition under which the optimal solution of \eqref{eq:tv-p} is consensual and identical to that of \eqref{eq:correct}.

\begin{theorem}\label{theorem1}
	Suppose that Assumptions \ref{ass:conv} and \ref{ass:lip} hold. If $\la\geq \lambda_0:=\max_{i\in\R}\|\n\E[F(\x^*,\xi_i)]\|_b$ with $p \geq 1$ and $b$ satisfying $\f{1}{b}+\f{1}{p}=1$, then we have $x^*=[\x^*]$, where $\x^*$ and $x^*$ are the optimal solutions of \eqref{eq:correct} and \eqref{eq:tv-p}, respectively.
\end{theorem}

Theorem \ref{theorem1} asserts that if the penalty constant $\la$ is selected to be large enough, the optimal solution of the regularized problem \eqref{eq:tv-p} is the same as that of \eqref{eq:correct}.
%The implication of Theorem \ref{theorem1} is that with a proper choice of $\la$, we can equivalently tackle \eqref{eq:tv-p}.
Next, we shall check the convergence properties of the RSA iterates with respect to the optimal solution of \eqref{eq:tv-p} under Byzantine attacks.

\begin{theorem}\label{theorem2}
	Suppose that Assumptions \ref{ass:conv}, \ref{ass:lip} and \ref{ass:bound} hold. Set the step size of $\ell_p$-norm RSA ($p\geq 1$) as $\A^{k+1} = \min\{\underline{\alpha},\frac{\overline{\alpha}}{k+1}\}$, where $\underline{\alpha}$ and $\overline{\alpha}$ depend on $\{\mu_0, \mu_i, L_0, L_i\}$. Then, for $k_0$ satisfying $\min\{k:\underline{\alpha} \geq \frac{\overline{\alpha}}{k+1}\}$, we have:
	\begin{equation}\label{theorem2_new_2-main}
	\!\E\|x^{k+1}-x^*\|^2\! \leq\!  (1-\eta \underline{\alpha})^{k}\|x^0-x^*\|^2+ \frac{\underline{\alpha} \Delta_0+ \Delta_2}{\eta},~ k < k_0
	\end{equation}
	and
	\begin{equation}\label{theorem2_10-main}
	\!\E\|x^{k+1}-x^*\|^2\leq \frac{\Delta_1}{k+1}+ \overline{\alpha}\Delta_2,~ k \geq k_0
	\end{equation}
	where $\eta$, $\Delta_1$ and $\D ={\cal O}(\la^2q^2)$ are certain positive constants.
\end{theorem}

Theorem \ref{theorem2} shows that the sequence of local iterates converge sublinearly to the near-optimal solution of the regularized problem \eqref{eq:tv-p}. The asymptotic sub-optimality gap is quadratically dependent on the number of Byzantine workers $q$. Building upon Theorems \ref{theorem1} and \ref{theorem2}, we can arrive at the following theorem.

%In other words, different from existing gradient aggregation methods \cite{Krum,geomatric,genneral-BSGD,trimmed_mean,Zeno} that have an explicit threshold on the tolerable number of %Byzantine workers, RSA can afford arbitrarily number of Byzantine workers though at the price of an enlarged sub-optimality gap.

\begin{theorem}\label{theorem3}
	Under the same assumptions as those in Theorem \ref{theorem2}, if we choose $\lambda \geq \lambda_0$ according to Theorem \ref{theorem1}, then for a sufficiently large $k \geq k_0$, we have:
	\begin{align}\label{eq:the3}
	\mathbb{E}||x^k-[\tilde{x}^*]||^2 \leq\frac{\Delta_1}{k+1} + \overline{\alpha}\Delta_2
	\end{align}
	If we choose $0<\lambda<\lambda_0$, and suppose that the difference between the optimizer of \eqref{eq:tv-p} and that of \eqref{eq:correct} is bounded by $||x^*-[\tilde{x}^*]||^2\leq\Delta_3$, then for $k \geq k_0$ we have:
	\begin{align}\label{eq:the3-1}
	\mathbb{E}||x^k-[\tilde{x}^*]||^2\leq\frac{2\Delta_1}{k+1}+2\overline{\alpha}\Delta_2+2\Delta_3
	\end{align}
\end{theorem}
Theorem \ref{theorem3} implies that the sequence of local iterates also converge sublinearly to the near-optimal solution of the original \eqref{eq:correct}. Under a properly selected $\la$, the sub-optimality gap in the limit is proportional to the number of Byzantine workers. Note that since the ${\cal O}(1/k)$ step size is quite sensitive to its initial value \cite{nemirovski2009}, we use the ${\cal O}(1/\sqrt{k})$ step size in our numerical tests. Its corresponding theoretical claim and convergence analysis are given in the supplementary document.

Regarding the optimal selection of the penalty constant $\la$ and the $\ell_p$ norm, a remark follows next.

\begin{remark}[Optimal selection of $\la$ and $p$]\label{remark 2}
Selecting different penalty constant $\la$ and $\ell_p$ norms in RSA generally leads to distinct performance. For a fixed $\la$, if a norm $\ell_p$ with a small $p$ is used, the dual norm $\ell_b$ has a large $b$ and thus results in a small $\lambda_0$ in Theorem \ref{theorem1}. Therefore, the local solutions are likely to be consensual. From the numerical tests, RSA with $\ell_{\infty}$ norm does not provide competitive performance, while those with $\ell_1$ and $\ell_2$ norms work well. On the other hand, for a fixed $p$, a small $\la$ cannot guarantee consensus among local solutions, but it gives a small sub-optimality gap $\Delta_2$. We recommend to use a $\la$ that is relatively smaller than $\lambda_0$, slightly sacrificing consensus but reducing the sub-optimality gap.
%However, if $p$ is small, for the same $\lambda$, the error introduced by the regularization term \eqref{eq:tv-p} can be large.
%On the other hand, for a fixed $p$, a small $\la$ cannot guarantee consensus among local solutions, but it gives a small sub-optimality gap $\Delta_2$ in Theorem \ref{theorem3}.
%However, if the gradients $\nabla \E[F(\tilde{x}^*,\epsilon_i)]$ deviations to the diagonal lines, we advise using smaller $p$. If the gradients $\nabla \E[F(\tilde{x}^*,\epsilon_i)]$ deviations to the axis lines, we advise using larger $p$.
\end{remark}

%\clearpage
\section{Numerical Tests}\label{sec.num}
In this section, we evaluate the robustness of the proposed RSA methods to Byzantine attacks and compare them with several benchmark algorithms. We conduct experiments on the MNIST dataset, which has 60k training samples and 10k testing samples, and use softmax regression with an $\ell_2$-norm regularization term $f_0(\tilde{x}) = \frac{0.01}{2}\|\tilde{x}\|^2$. We launch $20$ worker processes and $1$ master process on a computer with Intel i7-6700 CPU @ 3.40GHz. In the i.i.d. case, the training samples are randomly evenly assigned to the workers. In the heterogeneous case, every two workers evenly share the training samples of one digit. At every iteration, every regular worker estimates its local gradient on a mini-batch of 32 samples. The top-1 accuracy (evaluated with $x_0$ in RSA and $\tilde{x}$ in the benchmark algorithms) on the test dataset is used as the performance metric. { The code is available at https://github.com/liepill/rsa-byzantine}

\subsection{Benchmark algorithms}
We use the SGD iteration \eqref{eq:update:x} without attacks as the oracle, which is referred as \textbf{Ideal SGD}. Note that this method is not affected by $q$, the number of Byzantine workers. The other benchmark algorithms implement the following stochastic gradient aggregation recursion:
\begin{align}\label{eq:aggsgd}
\tilde{x}^{k+1} = \tilde{x}^k - \alpha^{k+1} \tilde{\nabla}(\tilde{x}^k)
\end{align}
where $\tilde{\nabla}(\tilde{x}^k)$ is an algorithm-dependent aggregated stochastic gradient that approximates the gradient direction, at the point $\tilde{x}^k$ sent by the master to the workers. Let the message sent by worker $i$ to the master be $v_i^k$, which is a stochastic gradient $\nabla F(\tilde{x}^k,\xi_i^k)$ if $i$ is regular, while arbitrary if $i$ is Byzantine. The benchmark algorithms use different rules to calculate the aggregated stochastic gradient.

\noindent \textbf{GeoMed} \cite{geomatric}.
The geometric median of $\{v_i^k:i\in[m]\}$ is denoted by:
\begin{align}
{\rm GeoMed}(\{v_i^k\})=\mathop{\arg\min}_{v\in\mathbb{R}^d}\sum_{i=1}^m||v-v_i^k||_2.
\end{align}
We use a fast Weiszfeld's algorithm \cite{weiszfeld2009point} to compute the geometric median in the experiments.

\noindent \textbf{Krum} \cite{Krum}. Krum calculates $\tilde{\nabla}(\tilde{x}^k)$ by:
\begin{align}\label{eq:krum}
{\rm Krum}(\{v_i^k\})=v_{i^*}^k, ~ i^*=\mathop{\arg\min}_{i\in[m]}\sum_{i\to j}||v_i^k-v_j^k||^2
\end{align}
where $i\to j (i\not= j)$ selects the indexes $j$ of the $m-q-2$ nearest neighbors of $v_i^k$ in $\{v_j^k: j\in[m]\}$, measured by Euclidean distances. Note that $q$, the number of Byzantine workers, must be known in advance in Krum.

\begin{figure}[h]
	\vspace{-0.2cm}
	\begin{center}
		\includegraphics[width=4.8cm]{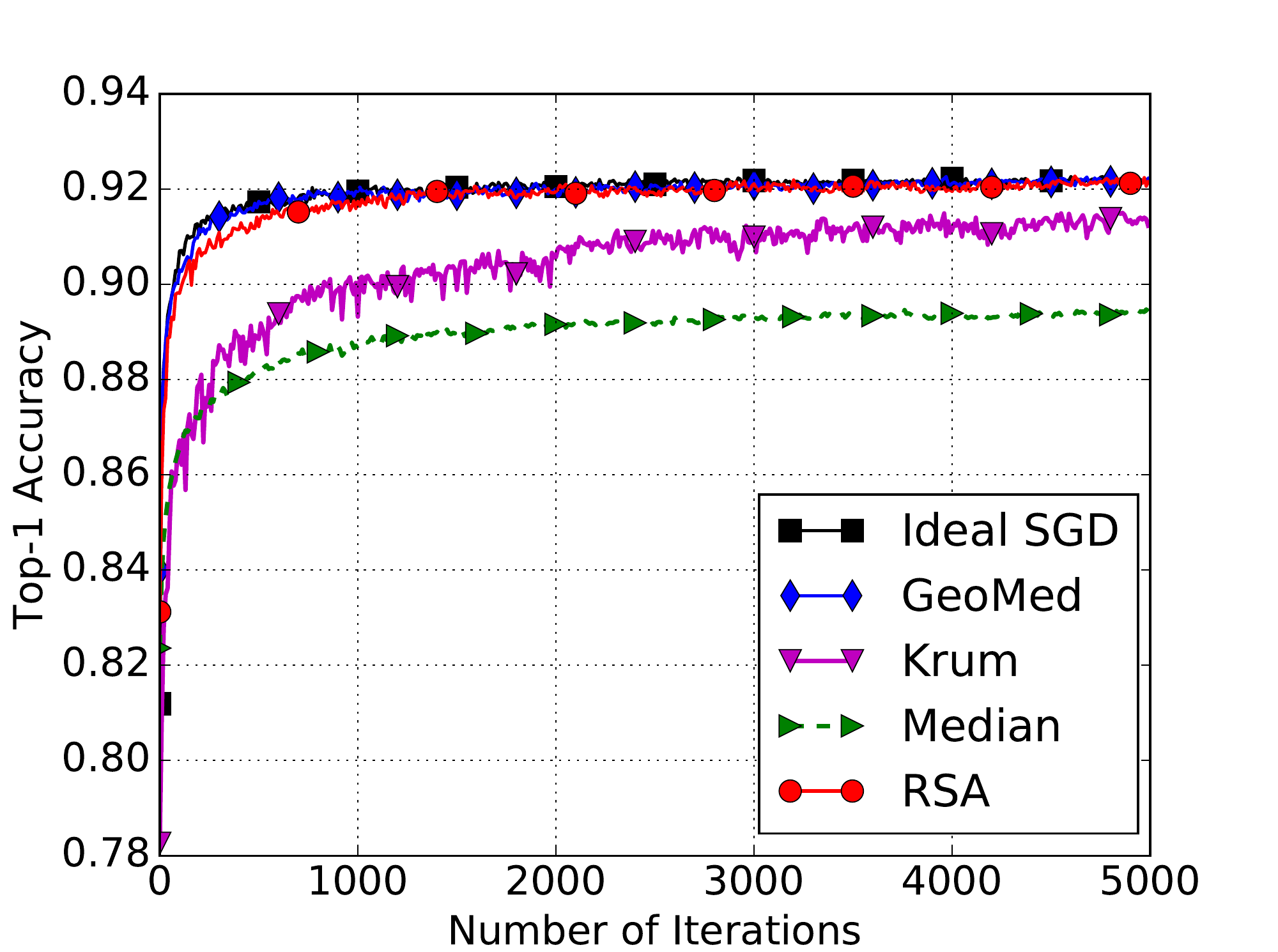}
		\vspace{-0.2cm}
		\caption{Top-1 accuracy without Byzantine attacks.}\label{eps:no}
		\vspace{-0.2cm}
	\end{center}
\end{figure}

\vspace{-0.2cm}

\noindent \textbf{Median} \cite{genneral-BSGD}. The marginal median aggregation rule returns the element-wise median of the vectors $\{v_i^k:i\in[m]\}$.

\noindent \textbf{SGD} \cite{bottou2010}. The classical SGD aggregates $\{v_i^k:i\in[m]\}$ by returning the mean, and is hence not robust to Byzantine attacks.

In the following experiments, step sizes of the benchmark algorithms are all hand-tuned to the best.

\subsection{Without Byzantine attacks}
In this test, we consider learning without Byzantine workers, and show the performance of all algorithms in Figure \ref{eps:no}. $\ell_1$-norm RSA chooses the parameter $\lambda = 0.1$ and the step size $\alpha^k=0.003/\sqrt{k}$. RSA and GeoMed are close to Ideal SGD, and significantly outperform Krum and Median. Therefore, robustifying the cost in RSA, though introduces bias, does not sacrifice performance in the regular case.

\subsection{Same-value attacks}
\begin{figure}[h]
	\hspace{-0.2cm}
	\begin{tabular}{cccc}
%		\hspace*{-3ex}
%		\includegraphics[width=4.8cm]{same_compare_q4_lam0.07_alpha0.001.eps}&
%		\hspace*{-5ex}
%		\includegraphics[width=4.8cm]{same_compare_q8_lam0.07_alpha0.001.eps}
%		\\
		\hspace*{-3ex}
		\includegraphics[width=4.8cm]{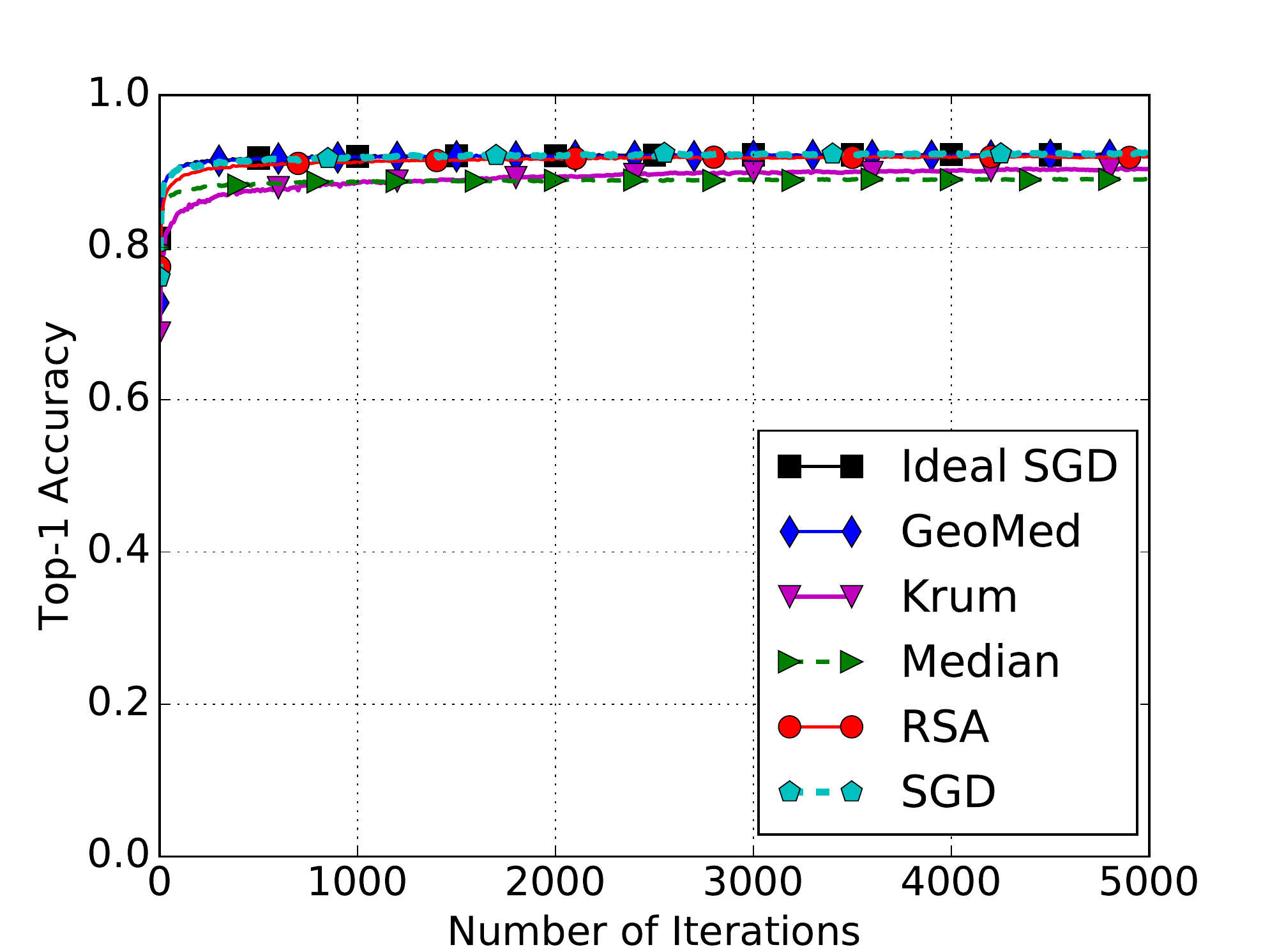}&
		\hspace*{-5ex}
		\includegraphics[width=4.8cm]{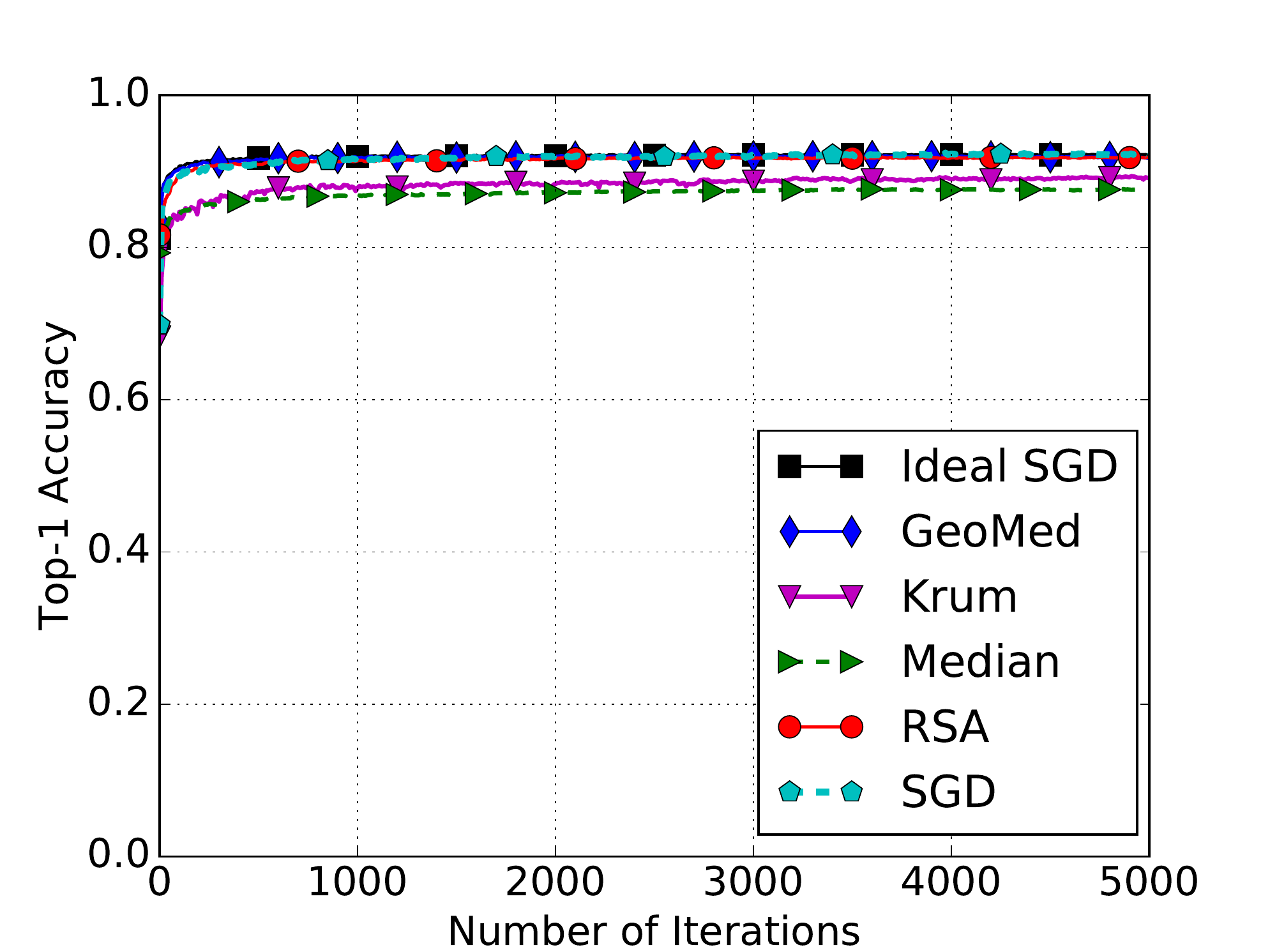}
		\\
		(a)& \hspace*{-5ex}(b)
	\end{tabular}
	\vspace{-0.2cm}
	\caption{{Top-1 accuracy under same-value attacks: (a) $q=4$ and $c=100$; (b) $q=8$ and $c=100$.}}\label{eps:same}
	\vspace{-0.2cm}
\end{figure}

{The same-value attacks set the message sent by a Byzantine worker $i$ as $v_i^k=c \textbf{1}$. Here $\textbf{1} \in\mathbb{R}^d$ is an all-one vector and $c$ is a constant, which we set as $100$. We consider two different numbers of Byzantine workers, $q=4$ and $q=8$, and demonstrate the performance in Figure \ref{eps:same}. $\ell_1$-norm RSA chooses the regularization parameter $\lambda=0.07$ and the step size $\alpha^k=0.001/\sqrt{k}$. When $q=4$, RSA and GeoMed are still close to Ideal SGD and outperform Krum and Median. When $q$ to $q=8$, Krum and Median perform worse than in $q=4$, while RSA and GeoMed are almost the same as in $q=4$. Note that due to the special structure of softmax regression, SGD still works under the same-value attacks.}
% We can see that under same-value attack, SGD still works, the reason is that in the \emph{softmax regression model}, there is a parameter redundancy problem. To be more clear,  the hypothesis estimates the calss label ptobabilities as\\
%$P(y^{(i)}=j|Data^{(i)};x)=\frac{e^{(x^j-\varphi)^T}Data^{(i)}}{\sum_{l=1}^{K} e^{(x^l-\varphi)^T Data^{(i)}}} =\frac{e^{(x^j)^T}Data^{(i)}}{\sum_{l=1}^{K} e^{(x^l)^T Data^{(i)}}} $, 
%\\ where $x \in \mathbb{R}^{K\times p}$ is the model parameter, $K$ is the number of data classes and $p$ is the number of features. $Data^{(i)}$ denotes the $i$-th data smaple and $y^{(i)}$ is the corresponding label. It's obviously that suntracting $\varphi$ from every $x^j$ does not affect our hypothesis predictions at all. So, when use SGD to solve softmax regression, same-value attack just like a fixed $\varphi$ and will not bias the model.} 
%When $q=4$, SGD fails, while RSA and GeoMed are still close to Ideal SGD and outperform Krum and Median. When $q$ is increased to $q=8$, Krum and Median perform worse than in $q=4$, while RSA and GeoMed are almost the same as in $q=4$.

\subsection{Sign-flipping attacks}
\begin{figure}[h]
	\vspace{-0.2cm}
	\hspace{-0.2cm}
	\begin{tabular}{cccc}
%		\hspace*{-3ex}
%		\includegraphics[width=4.8cm]{sign4_compare_q4_lam0.07_alpha0.001.eps}&
%		\hspace*{-5ex}
%		\includegraphics[width=4.8cm]{sign4_compare_q8_lam0.01_alpha0.0003.eps}
%		\\
		\hspace*{-3ex}
		\includegraphics[width=4.8cm]{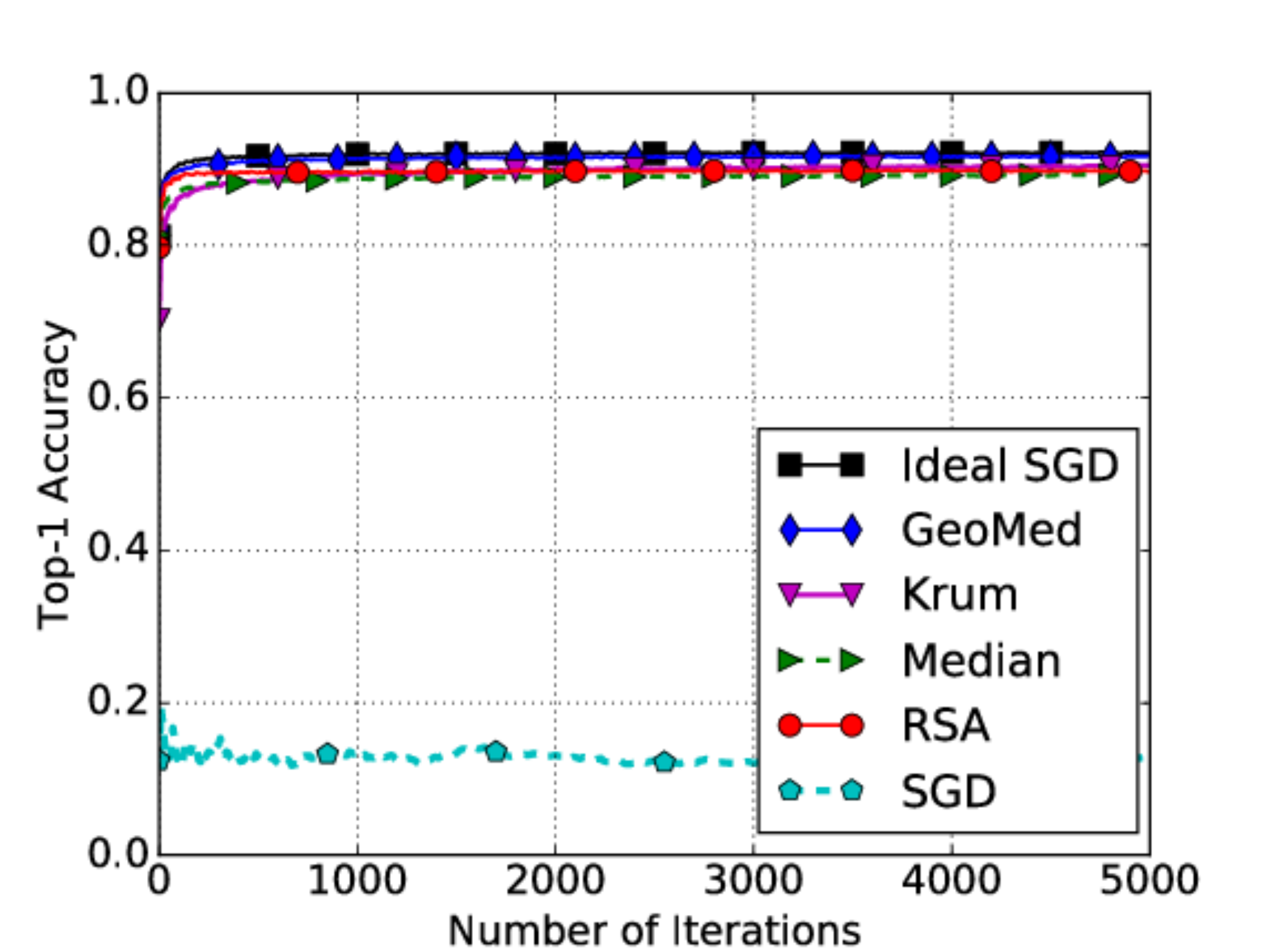}&
		\hspace*{-5ex}
		\includegraphics[width=4.8cm]{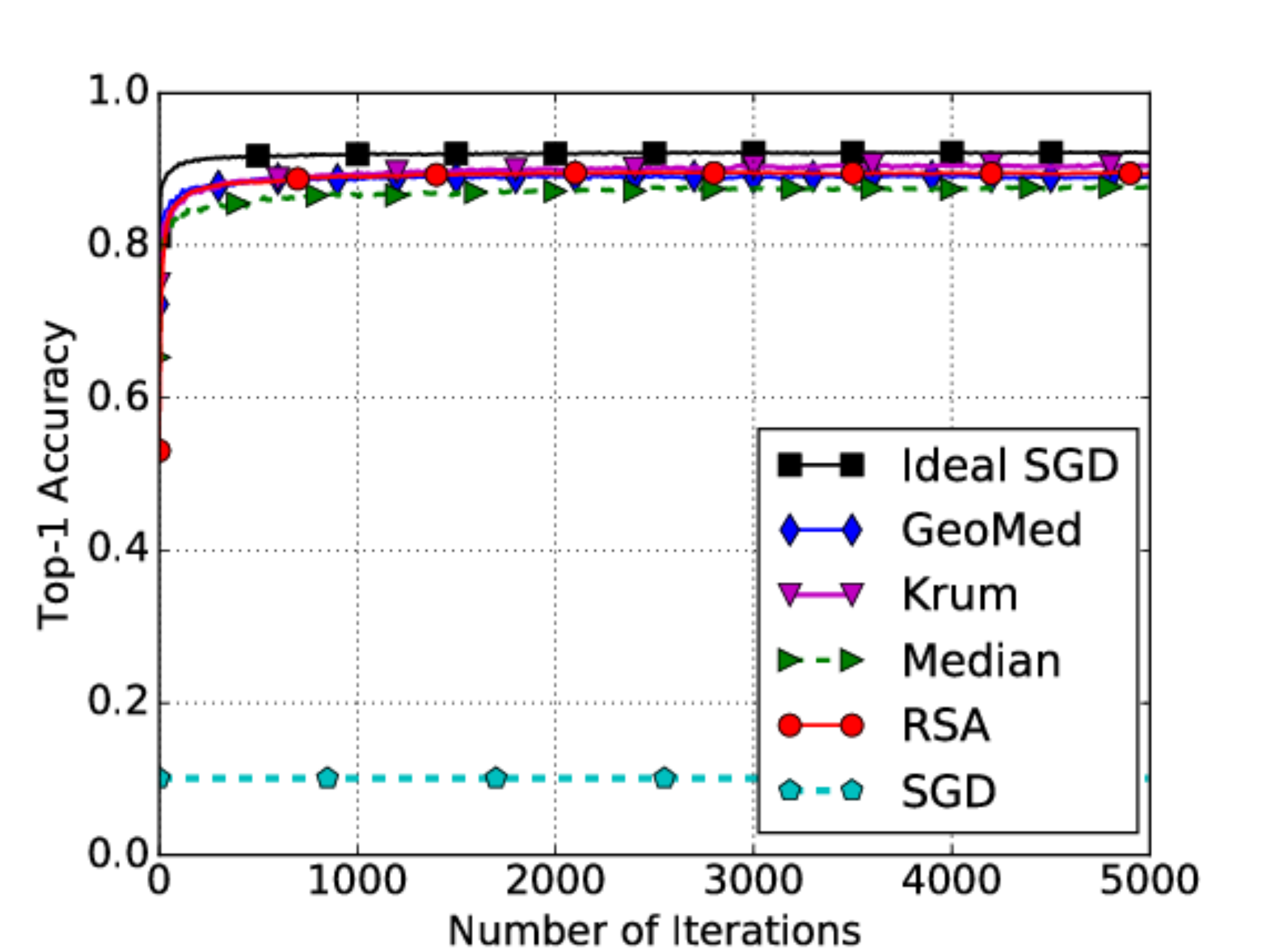}
		\\
		(a)&\hspace*{-5ex} (b)
	\end{tabular}
	\vspace{-0.2cm}
	\caption{Top-1 accuracy under sign-flipping attacks: (a) $q=4$ and $\sigma=-4$; (b) $q=8$ and $\sigma=-4$.}\label{eps:sign}
	\vspace{-0.2cm}
\end{figure}

The sign-flipping attacks flip the signs of messages (gradients or local iterates) and enlarge the magnitudes. To be specific, a Byzantine worker $i$ first calculates the true value $\hat{v}_i^k$, and then sends $v_i^k = \sigma \hat{v}_i^k$ to the master, where $\sigma$ is a negative constant. We test $\sigma=-4$ while set $q=4$ and $q=8$, as shown in Figure \ref{eps:sign}. The parameters are $\lambda=0.07$ and $\alpha=0.001/\sqrt{k}$ for $q=4$, while $\lambda=0.01$ and $\alpha=0.0003/\sqrt{k}$ for $q=8$. Not surprisingly, SGD fails in both cases. GeoMed, Median and $\ell_1$-norm RSA show similar performance, and Median is slightly worse than the other Byzantine-robust algorithms.

\subsection{Gaussian attacks}
\begin{figure}[h]
	\vspace{-0.2cm}
	\hspace{-0.2cm}
	\begin{tabular}{cccc}
		\hspace*{-3ex}
		\includegraphics[width=4.8cm]{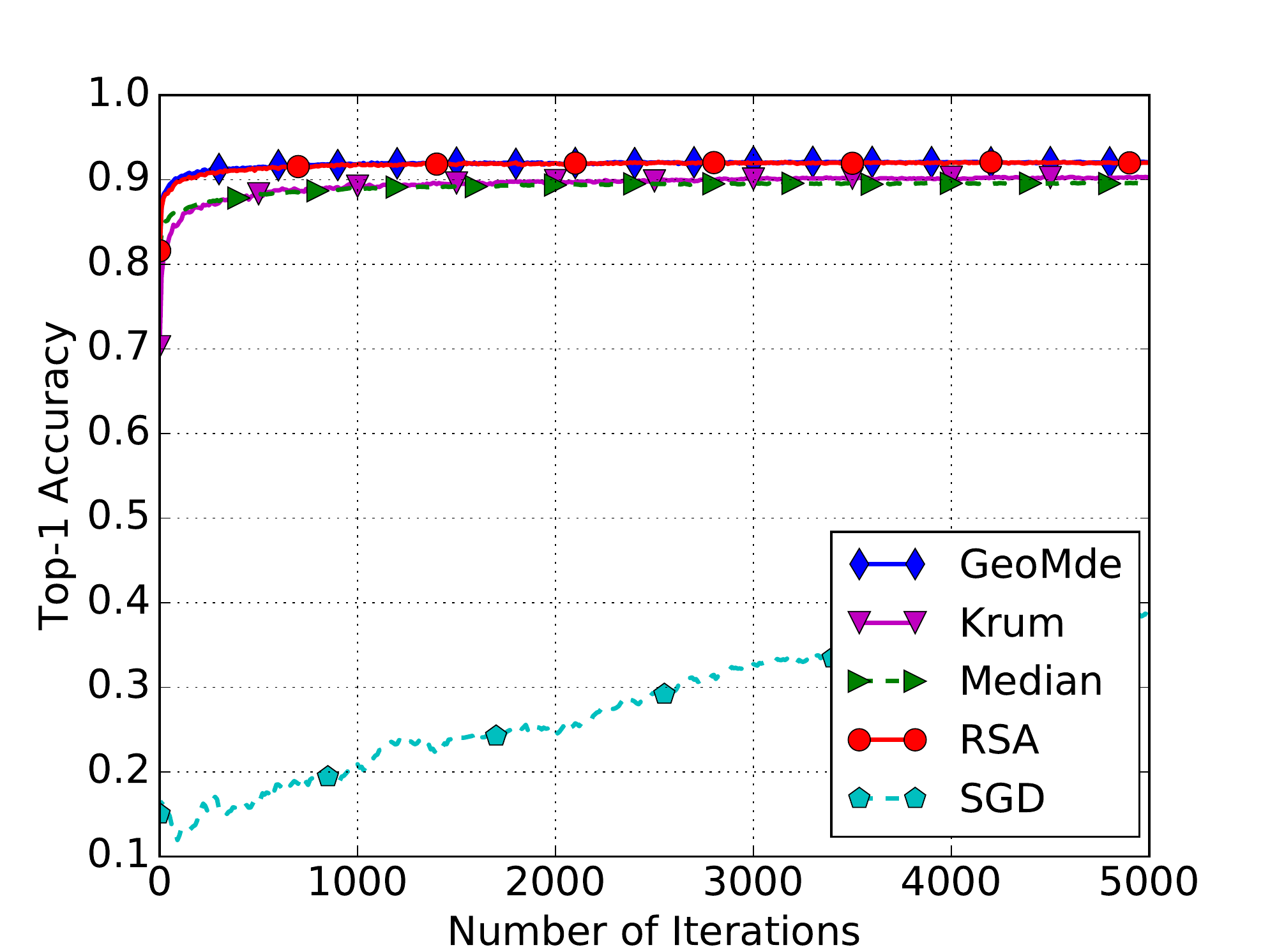}&
		\hspace*{-5ex}
		\includegraphics[width=4.8cm]{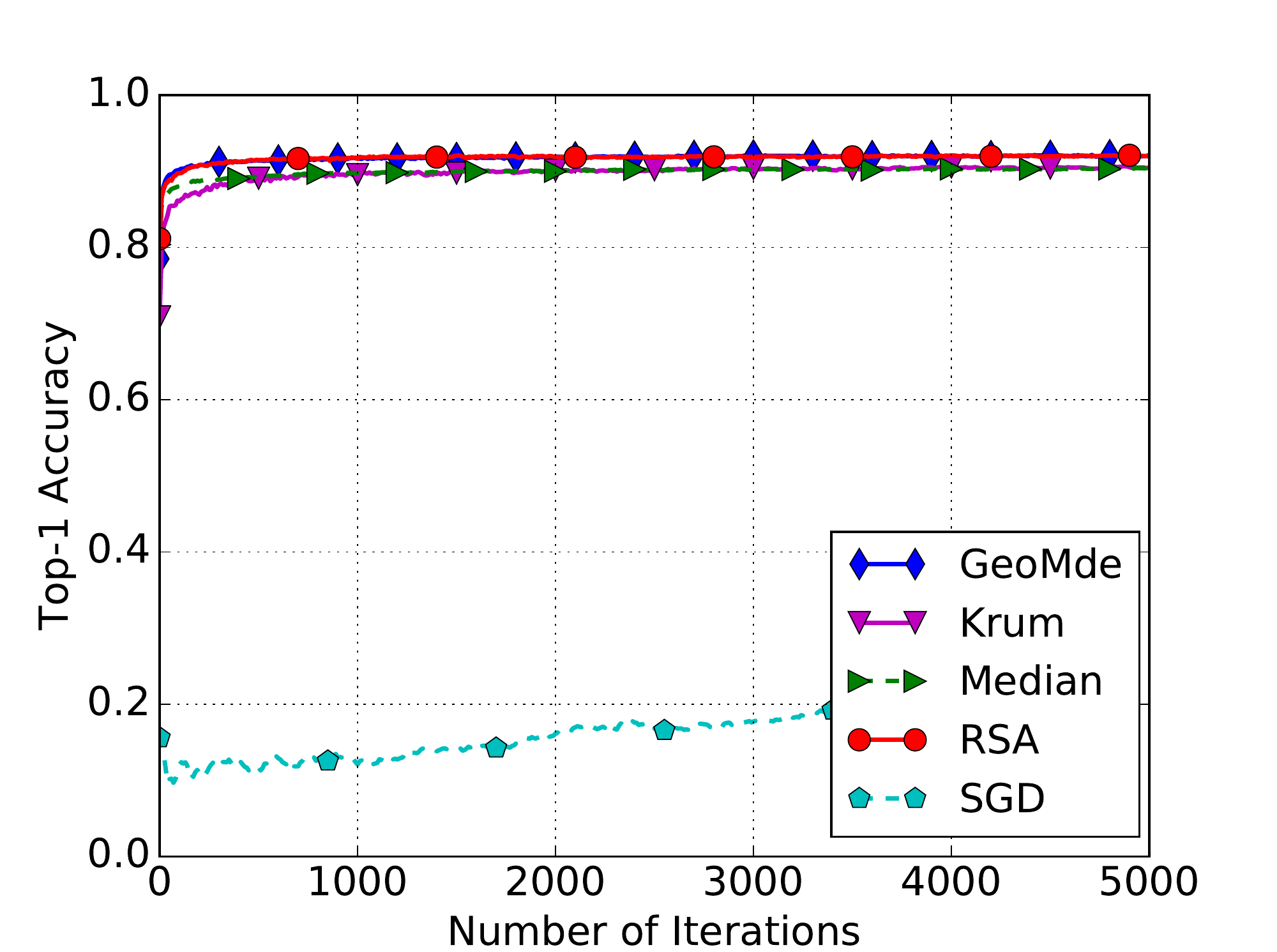}
		\\
		(a)&\hspace*{-5ex} (b)
	\end{tabular}
	\vspace{-0.2cm}
	\caption{{Top-1 accuracy under gaussian attacks: (a) $q=4$ and $c=10000$; (b) $q=8$ and $c=10000$.}}\label{eps:gaussian}
	\vspace{-0.2cm}
\end{figure}

 {The Gaussian attacks set the message sent by a Byzantine worker $i$ as $v_i^k$, where each element follows Gaussian distribution $\sim N(0,g^2)$. We test $g=10000$ while $q=4$ and $q=8$, as shown in Figure \ref{eps:gaussian}. In $\ell$-norm RSA, the parameters are $\lambda=0.07$ and $\alpha=0.002/\sqrt{k}$ for both $q=4$ and $q=8$. SGD falis in both cases. RSA and GeoMed have similar results and outperform Median and Krum.
% We consider the attackers that replace some of the messages (gradients or local iterates) with Gaussian random vectors with zero mean and isotropic covariance matrix with standard deviation 1.0 multiply a constant. To be specific, a Byzantine worker $i$
%sends message $v_i^k = c\xi_i^k$ to the master, $\xi_i^k \sim N(0, 1)$ and $c$ is a constant. We refer to this kind of attack as \emph{Gaussian attack}. We test $c=10000$ while set $q=4$ and $q=8$, as shown in Figure \ref{eps:gaussian}.
 }

\subsection{Runtime comparison}
\begin{figure}[h]
	\vspace{-0.2cm}
	\begin{center}
		\includegraphics[width=4.8cm]{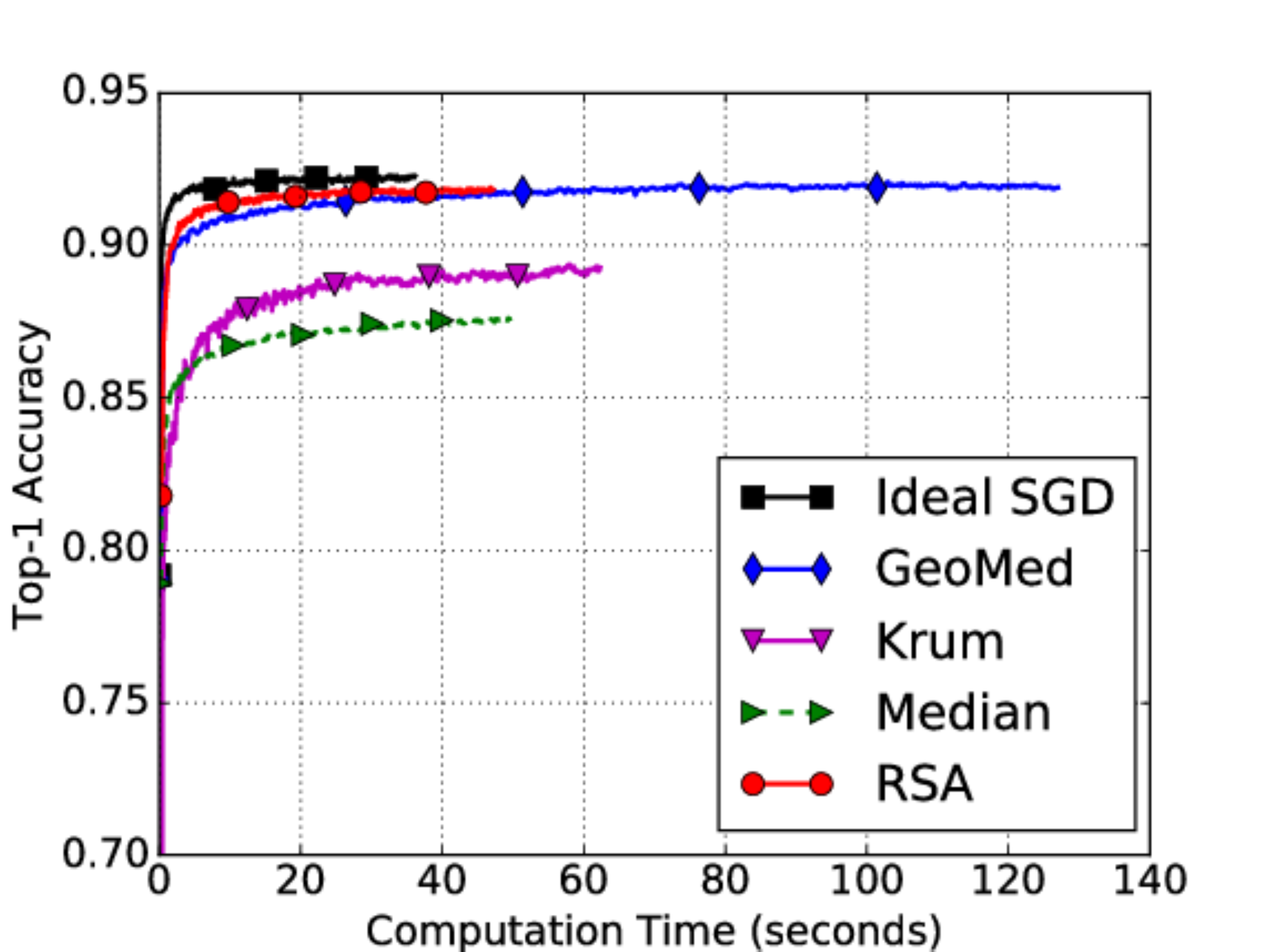}
		\vspace{-0.2cm}
		\caption{Runtime under same-value attacks with $q=8$ and $c=100$.}\label{eps:time}
		\vspace{-0.2cm}
	\end{center}
\end{figure}

%In this section, we compare the runtime of the several algorithms.
We show in Figure \ref{eps:time} the runtime of the algorithms under the same-value attacks with parameter $c=100$ and $q=8$ Byzantine workers. The total number of iterations for every algorithm is $5000$. Though the algorithms are not implemented in a federated learning platform, the comparison clearly demonstrates the additional per-iteration computational costs incurred in handling Byzantine attacks. GeoMed has the largest per-iteration computational cost due to the difficulty of calculating the geometric median. $\ell_1$-norm RSA and Median are both slightly slower than Ideal SGD, but faster than Krum. The only computational overhead of RSA than Ideal SDG lies in the computation of sign functions, which is light-weight. Therefore, RSA is advantageous in computational complexity comparing to other complicated gradient aggregation approaches.

\vspace{-0.2cm}

\vspace{-0.2cm}

\vspace{-0.2cm}

\vspace{0.7cm}
\subsection{Impact of hyper-parameter $\lambda$}
\begin{figure}[t]
	\centering
	\hspace{-0.2cm}
	\includegraphics[width=4.8cm]{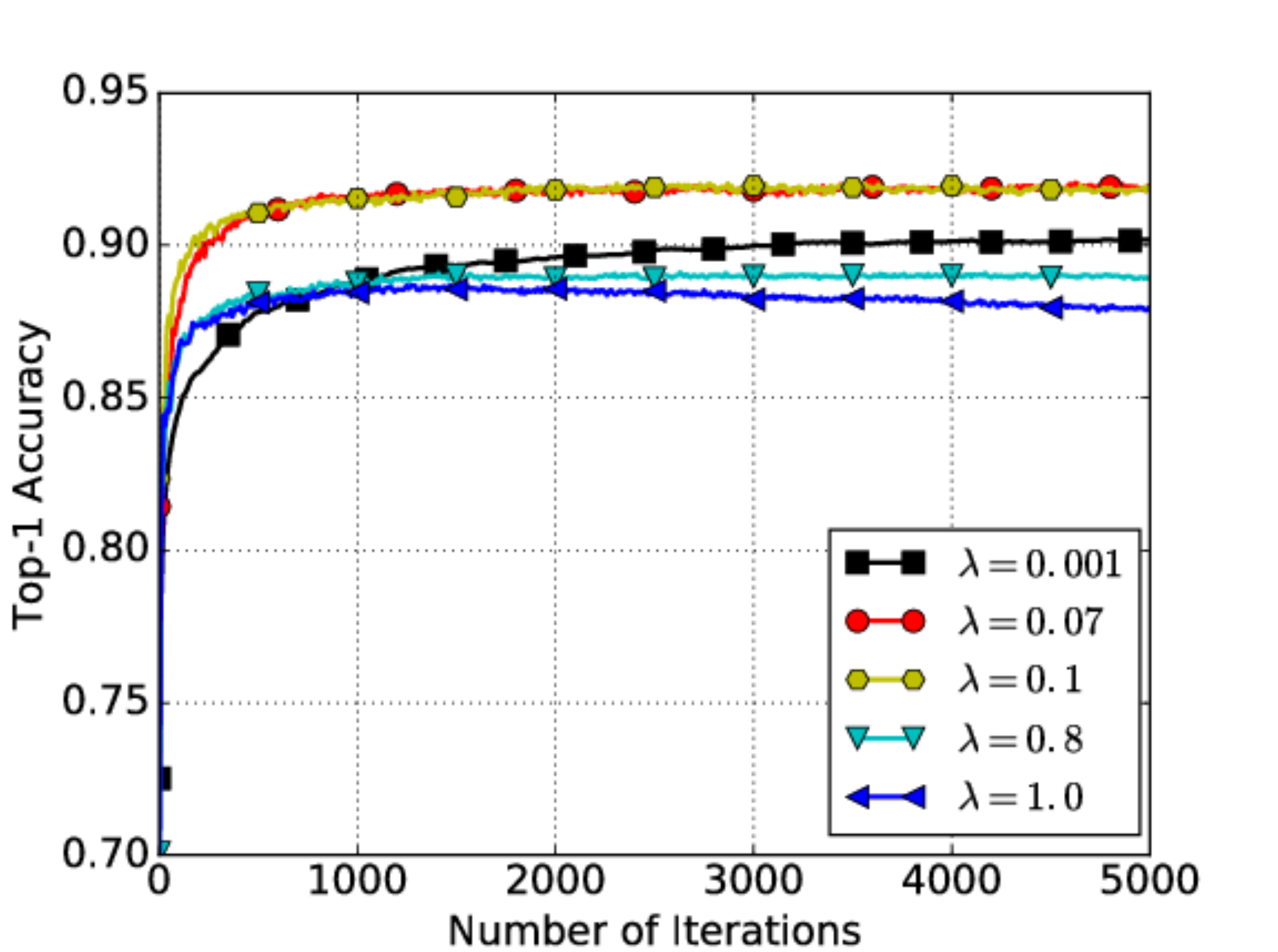}
	\vspace{-0.2cm}
	\caption{Top-1 accuracy with varying $\lambda$. We use same-value attacks with $q=8$ and $c=100$.}\label{eps:param}
	\vspace{-0.2cm}
\end{figure}

We vary the hyper-parameter $\lambda$, and show how it affects the performance. We use the same value attacks with $c=100$, vary $\lambda$, run RSA for $5000$ iterations, and depict the final top-1 accuracy in Figure \ref{eps:param}. The number of Byzantine workers is $q=8$ and the step sizes are hand-tuned to the best. Observe that when $\lambda$ is small, a regular worker tends to rely on its own data such that information fusion over the network is slow, which leads to slow convergence and large error. On the other hand, a large $\lambda$ also incurs remarkable error, as we have investigated in the convergence analysis.

	\vspace{-0.1cm}

\begin{figure}[h]
	\vspace{-0.2cm}
	\hspace{-0.2cm}
	\begin{tabular}{cccc}
		\hspace*{-3ex}
%		\includegraphics[width=4.8cm]{norm_compare_no_acc_lam0.1_alpha0.001.eps}&
%		\hspace*{-5ex}
%		\includegraphics[width=4.8cm]{norm_compare_no_var_lam0.1_alpha0.001.eps}
		\includegraphics[width=4.8cm]{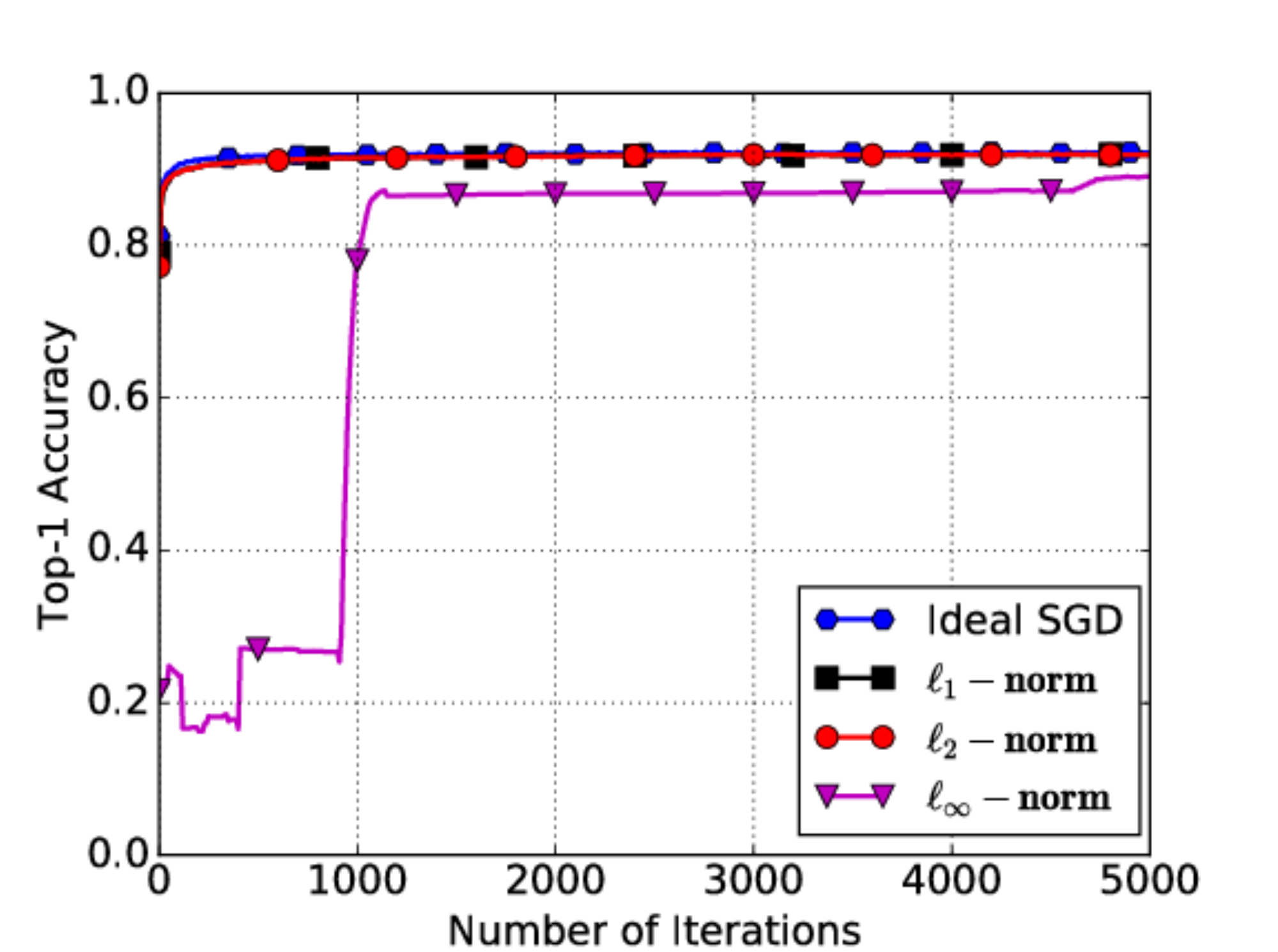}&
		\hspace*{-5ex}
		\includegraphics[width=4.8cm]{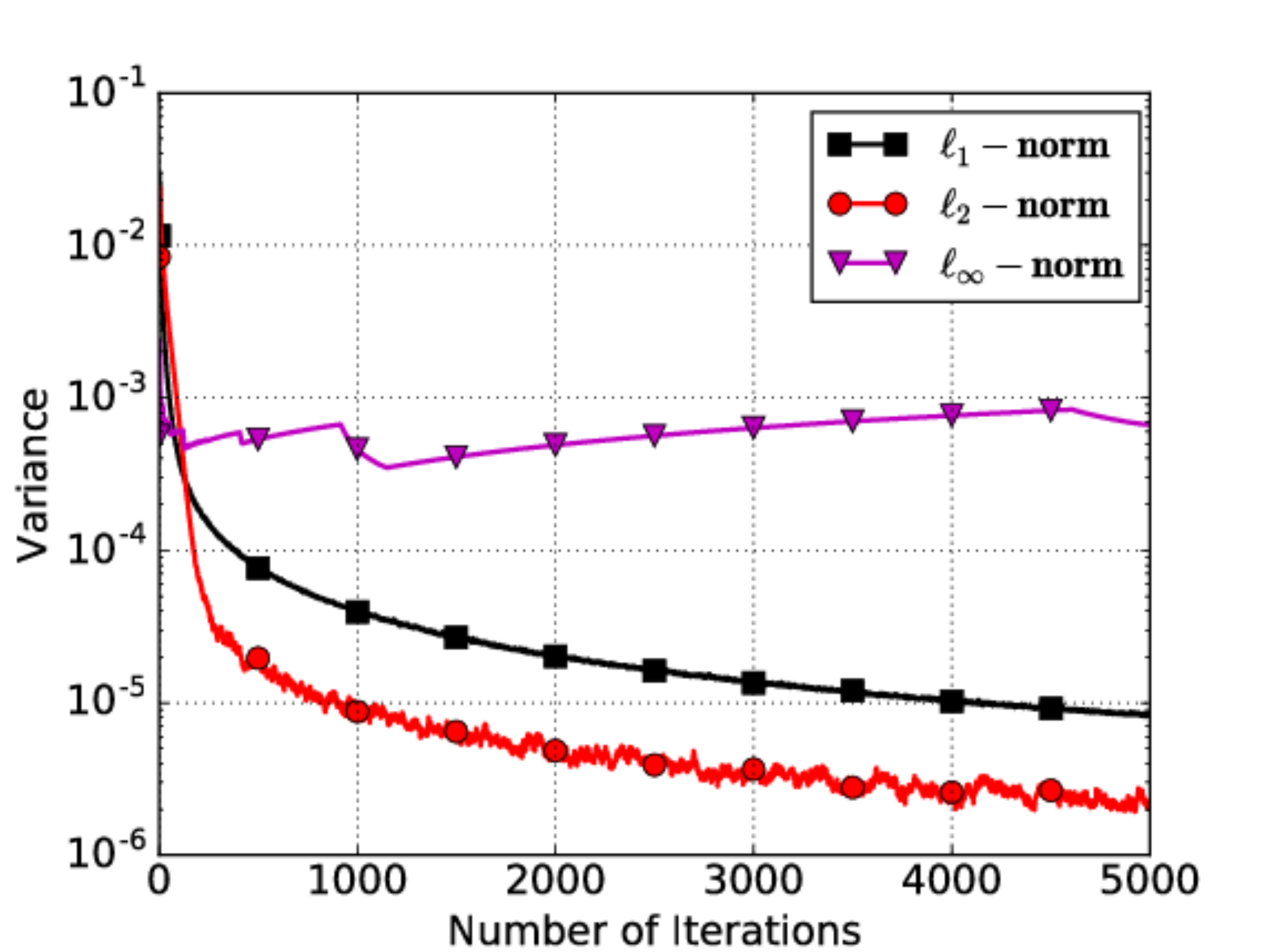}
		\\
		(a)& \hspace*{-5ex}(b)
	\end{tabular}
	\vspace{-0.2cm}
	\caption{RSA with different norms, without Byzantine attacks: (a) top-1 accuracy; (b) variance of regular workers' local iterates.}\label{eps:norm_no}
	\vspace{-0.2cm}
\end{figure}

	\vspace{-0.1cm}

\begin{figure}[h]
	%\vspace{-0.2cm}
	\hspace{-0.2cm}
	\begin{tabular}{cccc}
		\hspace*{-3ex}
%		\includegraphics[width=4.8cm]{norm_compare_acc_q8_lam0.07_alpha0.001_l21.2_0.001_max20_0.0001.eps}&
%		\hspace*{-5ex}
%		\includegraphics[width=4.8cm]{norm_compare_var_q8_lam0.07_alpha0.001_l21.2_0.001_max20_0.0001.eps}
		\includegraphics[width=4.8cm]{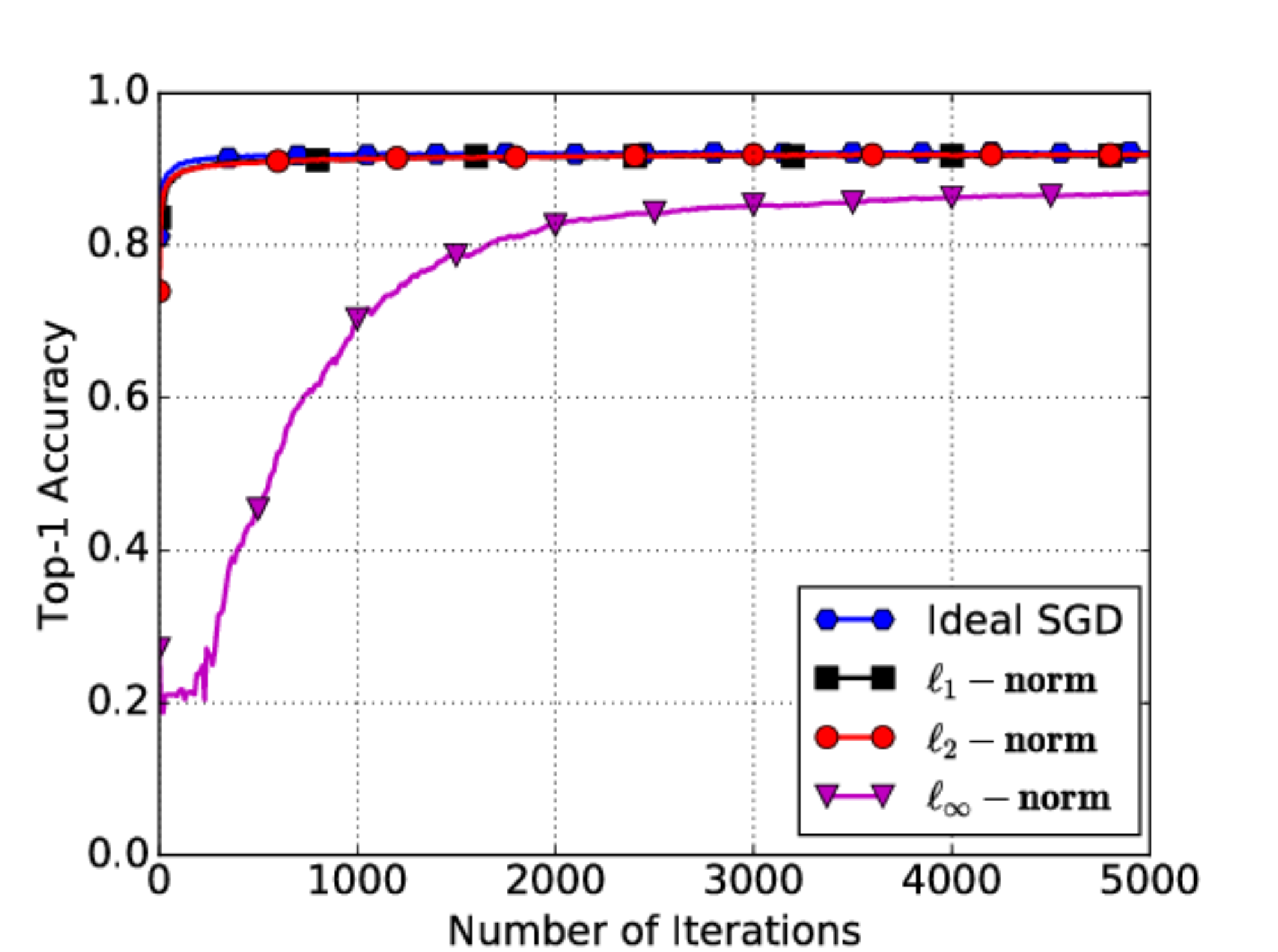}&
		\hspace*{-5ex}
		\includegraphics[width=4.8cm]{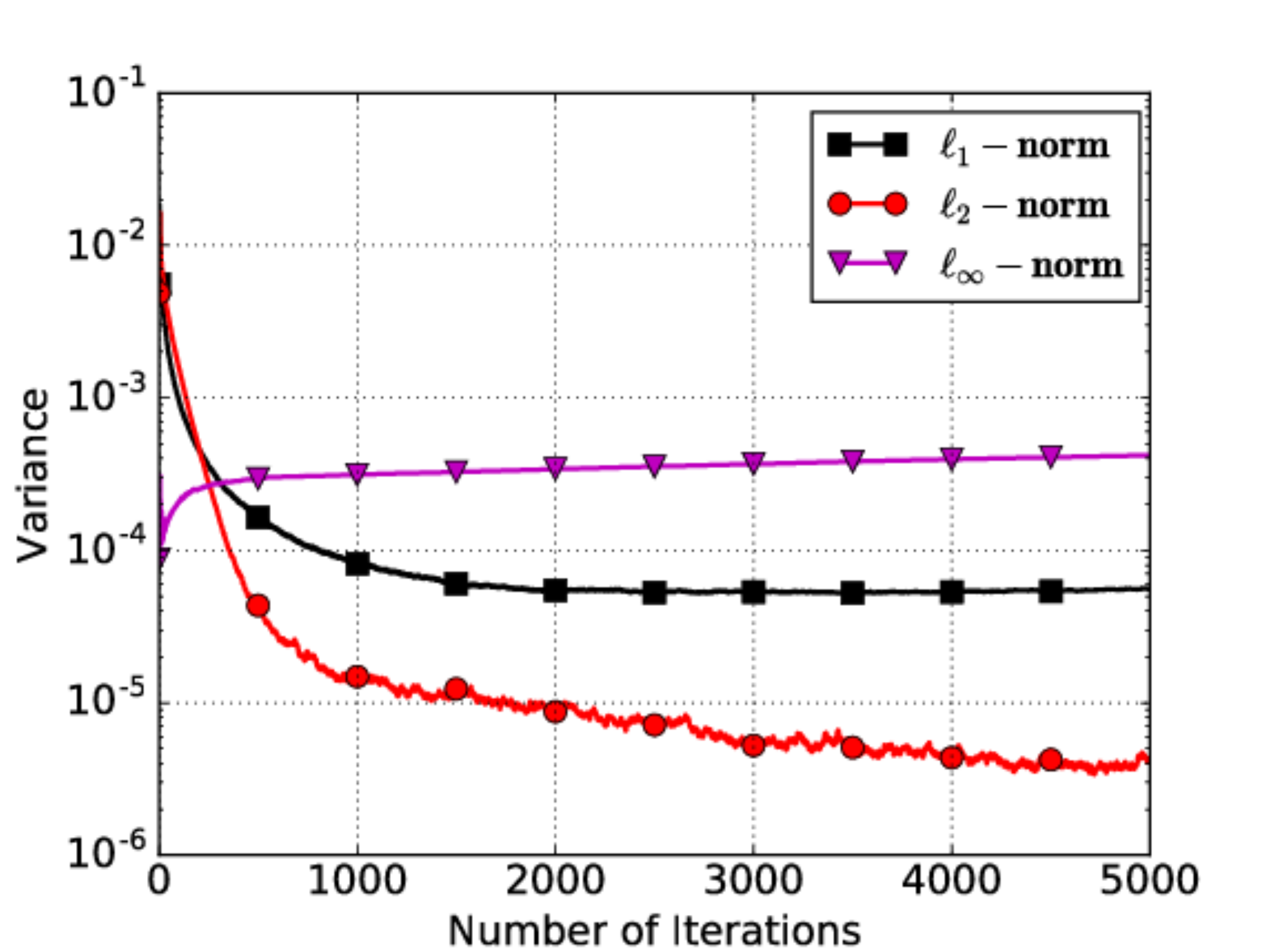}
		\\
		(a)& \hspace*{-5ex}(b)
	\end{tabular}
	\vspace{-0.2cm}
	\caption{RSA with different norms, under same-value attacks with $q=8$ and $c=100$: (a) top-1 accuracy; (b) variance of regular workers' local iterates.}\label{eps:norm_q8}
	\vspace{-0.2cm}
\end{figure}

\vspace{0.1cm}
\subsection{RSA with different norms}
\vspace{-0.1cm}
Now we compare RSA methods regularized with different norms. The results without Byzantine attacks and under the same-value attacks with $q=8$ and $c=100$ are demonstrated in Figures \ref{eps:norm_no} and \ref{eps:norm_q8}, respectively. We consider two performance metrics, top-1 accuracy and variance of the regular workers' local iterates. A small variance means that the regular workers reach similar solutions. Without the Byzantine attacks, $\ell_1$ is with $\lambda=0.1$ and $\alpha^k=0.001/\sqrt{k}$, $\ell_2$ is with $\lambda=1.4$ and $\alpha^k=0.001/\sqrt{k}$, while $\ell_\infty$ is with $\lambda=51$ and $\alpha^k=0.0001/\sqrt{k}$. Under the same-value attacks,  $\ell_1$ is with $\lambda=0.07$ and $\alpha^k=0.001/\sqrt{k}$, $\ell_2$ is with $\lambda=1.2$ and $\alpha^k=0.001/\sqrt{k}$, while $\ell_\infty$ is with $\lambda=20$ and $\alpha^k=0.0001/\sqrt{k}$. In both cases, $\ell_1$-norm RSA and $\ell_2$-norm RSA are close in terms of top-1 accuracy, and both of them is better than $\ell_\infty$-norm RSA. This observation coincides with our convergence analysis, namely, $\ell_\infty$-norm RSA needs a large $\lambda$ to ensure consensus, which in turn causes a large error. Indeed, we deliberately choose a not-too-large $\lambda$ for $\ell_\infty$-norm RSA so as to reduce the error, but sacrificing the consensus property. Therefore, regarding the variance of the regular workers' local iterates, $\ell_\infty$-norm RSA is the largest, while $\ell_2$-norm RSA is smaller than $\ell_1$-norm RSA.

%The parameters $\lambda$ and the step size $\alpha^k$ are hand-tuned to the best: without attacks, $\ell_1$ ($\lambda=0.1$ and $\alpha^k=0.001/\sqrt{k}$), $\ell_2$ ($\lambda=1.4$ %and $\alpha^k=0.001/\sqrt{k}$), $\ell_\infty$ ($\lambda=51$ and $\alpha^k=0.0001/\sqrt{k}$); under same-value attacks, $\ell_1$ ($\lambda=0.07$ and $\alpha^k=0.001/\sqrt{k}$), %$\ell_2$ ($\lambda=1.2$ and $\alpha^k=0.001/\sqrt{k}$), $\ell_\infty$ ($\lambda=20$ and $\alpha^k$ $=0.0001/\sqrt{k}$).

\subsection{Heterogeneous Data}
\begin{figure}[h!]
	\vspace{-0.2cm}
	\hspace{-0.2cm}
	\begin{tabular}{cccc}
		\hspace*{-3ex}
%		\includegraphics[width=4.8cm]{hete_compare_q4_lam0.5_alpha0.0005.eps}&
%		\hspace*{-5ex}
%		\includegraphics[width=4.8cm]{hete_compare_q8_lam0.5_alpha0.0005.eps}
		\includegraphics[width=4.8cm]{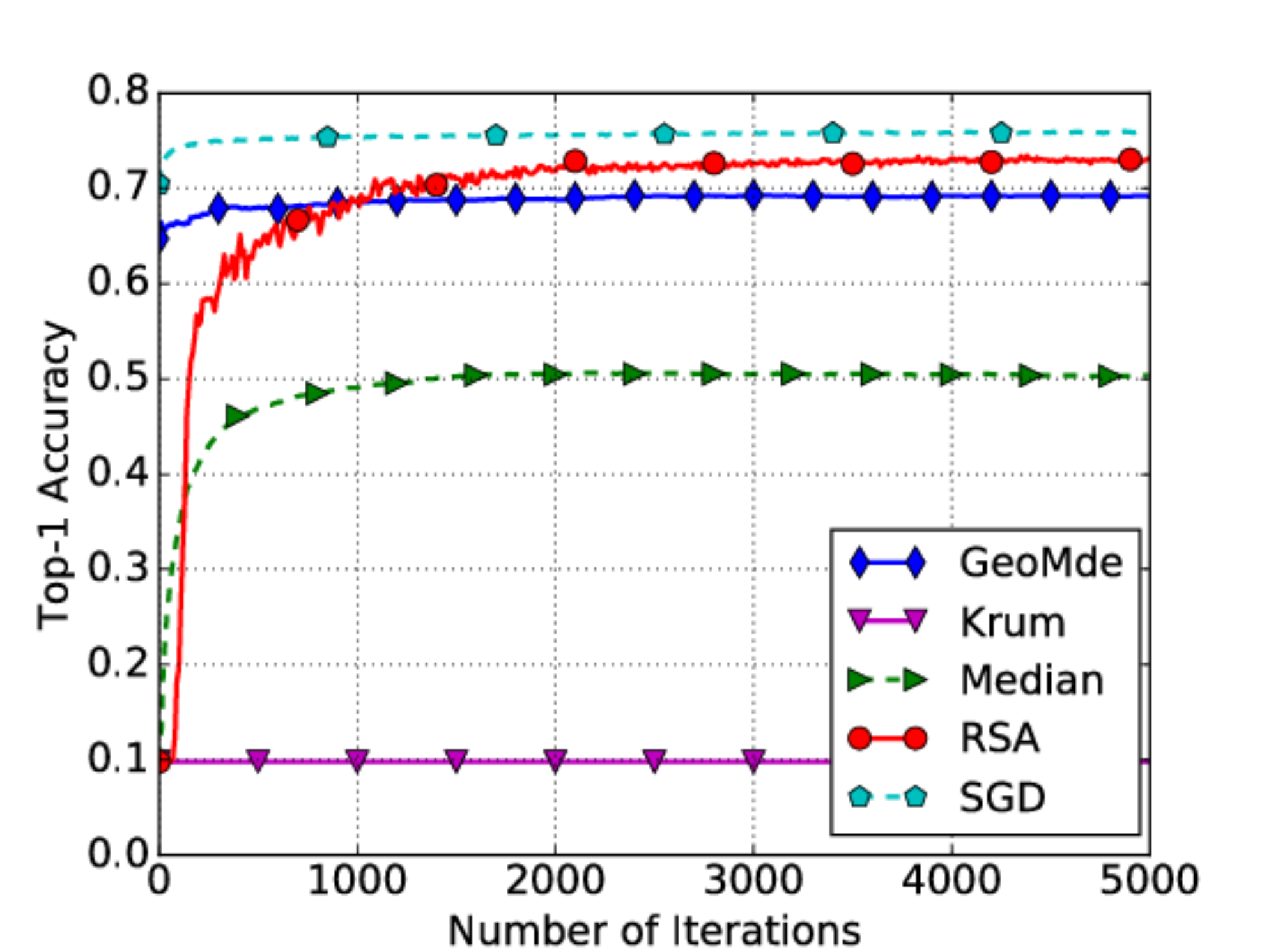}&
		\hspace*{-5ex}
		\includegraphics[width=4.8cm]{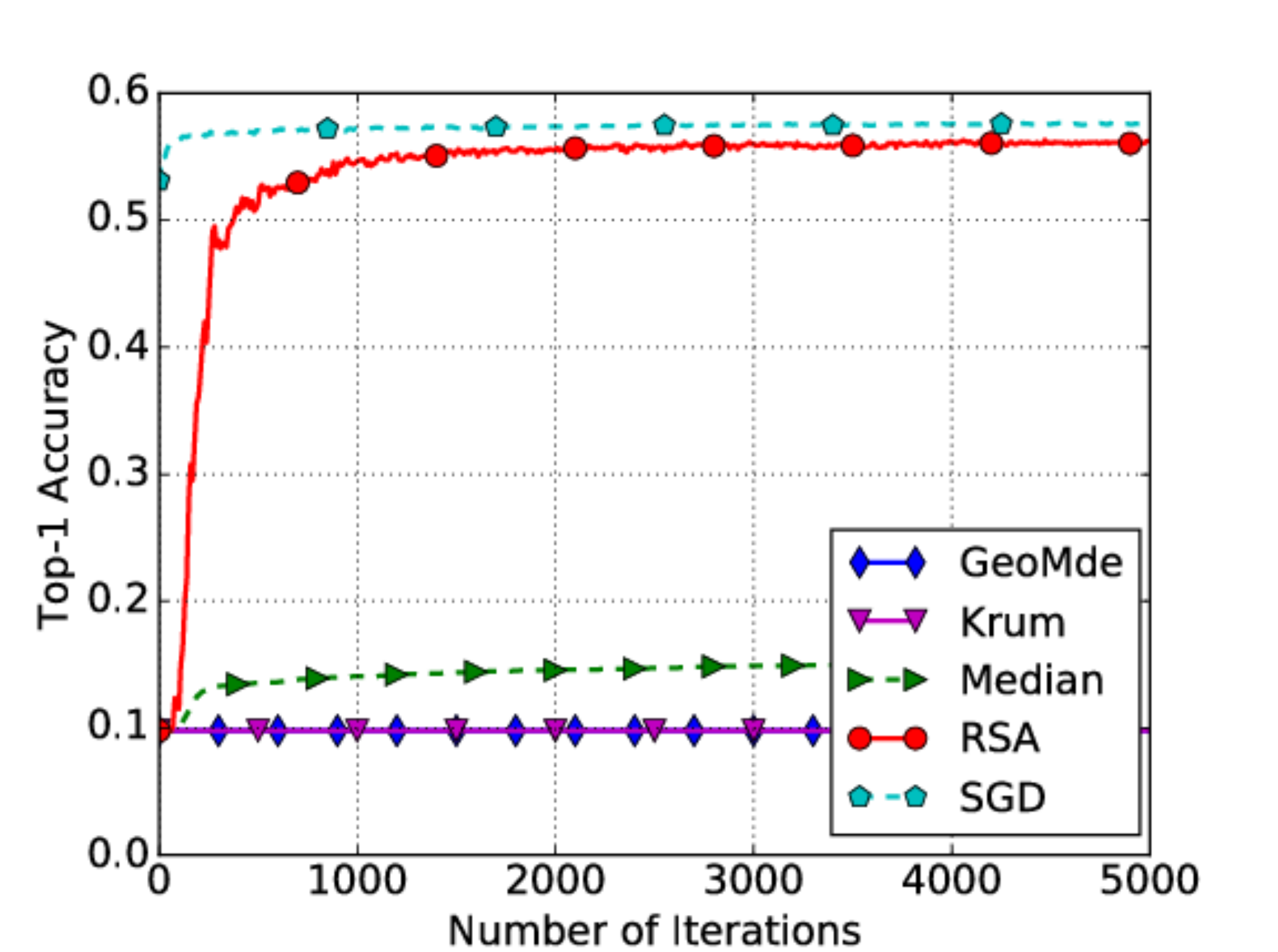}
		\\
		(a)& \hspace*{-5ex}(b)
	\end{tabular}
	\vspace{-0.2cm}
	\caption{Top-1 accuracy under attacks on heterogeneous data: (a) $q=4$; (b) $q=8$.}\label{eps:hete}
\end{figure}

To show the robustness of RSA on heterogeneous dataset, we re-distribute the MNIST data in this way: each two workers associate with the data about the same handwriting digit.
In experiment, each Byzantine worker $i$ transmits $v_i^k=v_r^k$, where worker $r$ is one of the regular workers. We set $r=1$ in the experiment. The results are shown in Figure \ref{eps:hete}. When $q=4$, two handwriting numbers' data are not available in the experiment, such that the best possible accuracy is around 0.8. When $q=8$, the best possible accuracy is around 0.6. The parameters of $\ell_1$-norm RSA are $\lambda=0.5$ and $\alpha^k=0.0005/\sqrt{k}$. Observe that when $q=4$, Krum fails, while RSA outperforms GeoMed and Median. When $q$ increases to 8, GeoMed, Krum and Median all fail, but RSA still performs well and reaches the near-optimal accuracy.

\section{Conclusions}
This paper dealt with distributed learning under Byzantine attacks. While the existing work mostly focuses on the case of i.i.d. data and relies on costly gradient aggregation rules, we developed an efficient variant of SGD for distributed learning from heterogeneous datasets under the Byzantine attacks.
{The resultant SGD-based methods that we term RSA converges to a near-optimal solution at an $O(1/k)$ convergence rate, where the optimality gap depends on the number of Byzantine workers. In the Byzantine-free settings, both SGD and RSA converge to the optimal solution at sublinear convergence rate.}
%The resultant subgradient-based algorithm termed RSA enjoys the sublinear convergence rate under Byzantine attacks, which is in the same order as SGD working in the Byzantine-free setting.
Numerically, experiments on real data corroborate the competitive performance of RSA compared to the state-of-the-art alternatives.

\subsubsection*{Acknowledgments.}
The work by T. Chen and G. Giannakis is supported in part by NSF 1500713 and 1711471. The work by Q. Ling is supported in part by NSF China 61573331 and Guangdong IIET 2017ZT07X355.

%\clearpage
\bibliographystyle{aaai}
\bibliography{aaairef}

\clearpage
\onecolumn
\appendix

\begin{center}
	{\Large \bf Supplementary Document for ``RSA: Byzantine-Robust Stochastic Aggregation Methods for \\ Distributed Learning from Heterogeneous Datasets''}
\end{center}

In this supplementary document, we present omitted proofs in the main manuscript.

\section{Proof of Proposition \ref{prop.1}}

\noindent \textit{Proof.} The proof has two parts.

\begin{itemize}
	\item Proof of $\{z\in \Rb^d: \La z,x\Ra=\|x\|_p,\ \|z\|_b\leq 1\} \subseteq \partial_x \|x\|_p$ :
	
	Considering any $z\in\{z\in \Rb^d: \La z,x\Ra=\|x\|_p,\ \|z\|_b\leq 1\}$, we have:
	\[
	\|x\|_p+\La z,y-x\Ra=\La z,y\Ra\leq\|z\|_b\|y\|_p\leq\|y\|_p
	\]
	where $\La z,y\Ra\leq \|z\|_b\|y\|_p$ due to Holder's inequality. According to the definition of subdifferential, it holds that $z\in\partial_x \|x\|_p$.
	
	\item Proof of $\partial_x \|x\|_p \subseteq \{z\in \Rb^d: \La z,x\Ra=\|x\|_p,\ \|z\|_b\leq 1\}$ :
	
	Considering any $z\in\partial_x \|x\|_p$, one can always find a vector $x_z$ that satisfies $\|x_z\|_{p}=1$ and $\La z,x_z\Ra=\|z\|_b$ since $\frac{1}{b}+\frac{1}{p} = 1$. Let $y=\|x\|_{p}x_z$, we have:
	\[
	\|y\|_p-\|x\|_p\geq \La z,y-x\Ra=\La z,y\Ra-\La z,x\Ra=\|x\|_p\|z\|_b-\La z,x\Ra
	\]
	where the first inequality comes from the definition of subdifferential. Since $\|y\|_p-\|x\|_p=0$ when $y=\|x\|_{p}x_z$ and $\|x_z\|_{p}=1$, the above result yields $0\geq\|x\|_p\|z\|_b-\La z,x\Ra$. However, due to the Holder's inequality it also holds $\|x\|_p\|z\|_b\geq\La z,x\Ra$. Thus, we must have $\|x\|_p\|z\|_b=\La z,x\Ra$.
	
	For $x\neq 0$, we use the definition of subdifferential to derive $\|2x\|_p-\|x\|_p\geq \La z,x\Ra$ and $\|0\|_p-\|x\|_p\geq\La z,-x\Ra$, from which we conclude that $\La z,x\Ra=\|x\|_p$. Since $\|x\|_p\|z\|_b=\La z,x\Ra$, we have $\|z\|_b=1$.
	
	For $x=0$, it holds that $\La z,x\Ra=\|x\|_p$. Due to $\|x_z\|_p-\|0\|_p\geq \La z,x_z\Ra$ from the definition of subdifferential, as well as $\|x_z\|_p=1$ and $\La z,x_z\Ra=\|z\|_b$ by hypothesis, we have $\|z\|_b\leq1$.
\end{itemize}

\section{Proof of Theorem \ref{theorem1}}

\noindent \textit{Proof.} Since $\la_0=\max_{i\in\R}\|\n\E[F(\x^*,\xi_i)]\|_b$ and $\la\geq\la_0$, we have $\n\E[F(\x^*,\xi_i)]\in \{\la z:\|z\|_b\leq 1\}$. As $p \geq 1$ and $\f{1}{b}+\f{1}{p}=1$, by Proposition \ref{prop.1}, we have $\n\E\big[F(\x^*,\xi_i)\big]\in \la\partial\|0\|_p$ for all $i\in \R$, and consequently:
\begin{equation}\label{pf.theorem1_2}
0\in\n\E\big[F(\x^*,\xi_i)\big]+\la\partial\|\x^*-\x^*\|_p, ~ \forall i \in \R
\end{equation}

Also, by $\n\E\big[F(\x^*,\xi_i)\big]\in \la\partial\|0\|_p$ for all $i\in \R$, there exists $\nu_i\in \partial\|0\|_p$ such that:
\begin{equation}\label{pf.theorem1_1}
\n\E\big[F(\x^*,\xi_i)\big]+\la \nu_i=0, ~ \forall i \in \R
\end{equation}
Summing \eqref{pf.theorem1_1} up for $i\in \R$, we have:
\begin{equation}\label{pf.theorem1_3}
\sum_{i\in\R}\bigg(\n\E\big[F(\x^*,\xi_i)\big]+\la \nu_i\bigg)=0
\end{equation}
From the optimality condition of \eqref{eq:correct}, $\sum_{i\in\R}\n\E\big[F(\x^*,\xi_i)\big]$ $+\n f_0(\x^*)=0$. Substituting this equality to \eqref{pf.theorem1_3}, we have:
\begin{equation}
\n f_0(\x^*)-\sum_{i\in\R} \la \nu_i=0
\end{equation}
Because every $\nu_i$ is a vector satisfying $\nu_i\in \partial\|0\|_p$, it is straightforward to conclude that:
\begin{equation}\label{pf.theorem1_4}
0\in \n f_0(\x^*)+\sum_{i\in\R} \la \partial\|\x^*-\x^*\|_p
\end{equation}

Combining \eqref{pf.theorem1_2} and \eqref{pf.theorem1_4}, we know that $x^* := [\x^*]$ (that is, $x_i^* = \x^*$ for all $i \in \R$ and $x_0^* = \x^*$) satisfies the optimality condition of \eqref{eq:tv-p}. This solution is also unique due to Assumption \ref{ass:conv}.

\section{Proof of Theorem \ref{theorem2}}

We first give a complete form of Theorem \ref{theorem2} as follows.

\vspace{0.5em}

\noindent \textbf{A complete form of Theorem \ref{theorem2}.} Suppose that Assumptions \ref{ass:conv}, \ref{ass:lip} and \ref{ass:bound} hold. Set the step size of $\ell_p$-norm RSA ($p\geq 1$) as $\A^{k+1} = \min\{\underline{\alpha},\frac{\overline{\alpha}}{k+1}\}$, where $\underline{\alpha} = \min\{ \min_{i \in \mathcal{R}} \frac{1}{2(\mu_i + L_i)}, \frac{1}{2(\mu_0 + L_0)} \}$, and $\overline{\alpha} > \frac{1}{\eta}$ with $\eta = \min\{ \min_{i \in \mathcal{R}} \frac{2\mu_i L_i}{\mu_i + L_i}, \frac{2\mu_0 L_0}{\mu_0 + L_0} - \epsilon \}$, and $\epsilon$ is any constant within $(0, \frac{2\mu_0 L_0}{\mu_0+L_0})$. Then, there exists a smallest integer $k_0$ satisfying $\underline{\alpha} \geq \frac{\overline{\alpha}}{k_0+1}$ such that:
\begin{equation}\label{theorem2_new_2-supp}
\E\|x^{k+1}-x^*\|^2 \leq (1-\eta \underline{\alpha})^{k} \E\|x^0-x^*\|^2+ \frac{1}{\eta} (\underline{\alpha} \Delta_0+ \Delta_2), \quad \forall k < k_0
\end{equation}
and
\begin{equation}\label{theorem2_10-supp}
\E\|x^{k+1}-x^*\|^2\leq \frac{\Delta_1}{k+1}+ \overline{\alpha}\Delta_2, \quad \forall k \geq k_0.
\end{equation}
Here we define
$$\Delta_1 = \max\left\{ \frac{\overline{\alpha}^2 \Delta_0}{\eta \overline{\alpha} -1}, (k_0+1) \E\|x^{k_0}-x^*\|^2 + \frac{\overline{\alpha}^2 \Delta_0}{k_0+1} \right\}$$
as well as
$$\Delta_0 = 16\la^2rd+16\la^2r^2d+2\la^2q^2d+2\sum_{i\in\R}\delta_i^2 \quad \text{and} \quad \Delta_2 = \f{\la^2q^2d}{\epsilon}.$$

\vspace{0.5em}

\noindent \textit{Proof.} The proof contains the following steps.

\noindent \textbf{Step 1.} From the RSA update \eqref{eq:update2:x:i} at every regular worker $i$, we have:
\begin{equation}\label{theorem2_1}
\begin{split}
 &\E\|x_i^{k+1}-x_i^*\|^2\\
=&\E\|x_i^k-x_i^*-\A^{k+1}\bigg(\nabla F(x_i^k,\xi_i^k)+\la\partial_{x_i} \|x_i^k-x_0^k\|_p\bigg)\|^2\\
=&\E\|x_i^k-x_i^*\|^2+(\A^{k+1})^2\E\|\nabla F(x_i^k,\xi_i^k)+\la \partial_{x_i} \|x_i^k-x_0^k\|_p\|^2\\
&-2\A^{k+1}\E\La\nabla F(x_i^k,\xi_i^k),x_i^k-x_i^*\Ra-2\A^{k+1}\E\La\la \partial_{x_i} \|x_i^k-x_0^k\|_p,x_i^k-x_i^*\Ra
\end{split}
\end{equation}
Since $x_i^k$ is independent with $\xi_i^k$, it follows that:
\begin{equation}\label{theorem2_2}
\E\big[\La\nabla F(x_i^k,\xi_i^k),x_i^k-x_i^*\Ra\big] = \E\big[ \La \nabla \E\big[F(x_i^k,\xi_i^k)\big], x_i^k-x_i^* \Ra\big].
\end{equation}
Substituting \eqref{theorem2_2} to \eqref{theorem2_1}, we have:
\begin{equation}\label{theorem2_3}
\begin{split}
\E\|x_i^{k+1}-x_i^*\|^2=&\E\|x_i^k-x_i^*\|^2+(\A^{k+1})^2\E\|\nabla F(x_i^k,\xi_i^k)+\la \partial_{x_i} \|x_i^k-x_0^k\|_p\|^2\\
&-2\A^{k+1}\E\La\nabla \E\big[F(x_i^k,\xi_i^k)\big],x_i^k-x_i^*\Ra-2\A^{k+1}\E\La\la \partial_{x_i} \|x_i^k-x_0^k\|_p,x_i^k-x_i^*\Ra\\
\stackrel{(a)}{=}&\E\|x_i^k-x_i^*\|^2+(\A^{k+1})^2\E\|\nabla F(x_i^k,\xi_i^k)+\la \partial_{x_i} \|x_i^k-x_0^k\|_p\|^2\\
&-2\A^{k+1}\E\La\nabla \E\big[F(x_i^k,\xi_i^k)\big]-\nabla \E\big[F(x_i^*,\xi_i^k)\big],x_i^k-x_i^*\Ra\\
&-2\A^{k+1}\E\La\la \partial_{x_i} \|x_i^k-x_0^k\|_p-\la \partial_{x_i} \|x_i^*-x_0^*\|_p,x_i^k-x_i^*\Ra
\end{split}
\end{equation}
where in (a), we insert the optimality condition of \eqref{eq:tv-p} with respect to $x_i$, namely, $\E\big[\nabla F(x_i^*,\xi_i)\big] + \la \partial_{x_i} \|x_i^*-x_0^*\|_p = 0$, and replace $\xi_i$ by $\xi_i^k$ due to $\xi_i^k \sim \mathcal{D}_i$.

For the second term at the right-hand side of \eqref{theorem2_3}, we have:
\begin{equation}\label{theorem2_3_2-tmp}
\begin{split}
&\E\|\nabla F(x_i^k,\xi_i^k)+\la \partial_{x_i} \|x_i^k-x_0^k\|_p\|^2\\
=   &\E\|\nabla \E\big[ F(x_i^k,\xi_i^k)\big]+\la \partial_{x_i} \|x_i^k-x_0^k\|_p+\n F(x_i^k,\xi_i^k)-\n\E\big[ F(x_i^k,\xi_i^k)\big]\|^2\\
\leq&2\E\|\nabla \E\big[ F(x_i^k,\xi_i^k)\big]+\la \partial_{x_i} \|x_i^k-x_0^k\|_p\|^2+2\E[\|\n F(x_i^k,\xi_i^k)-\n\E\big[ F(x_i^k,\xi_i^k)\big]\|^2]\\
\stackrel{(b)}{\leq}&2\E\|\nabla \E\big[ F(x_i^k,\xi_i^k)\big]+\la \partial_{x_i} \|x_i^k-x_0^k\|_p\|^2+2\delta_i^2
\end{split}
\end{equation}
where (b) is due to the bounded variance given by Assumption \ref{ass:bound}. Plugging the optimality condition $\E\big[\nabla F(x_i^*,\xi_i^k)\big] + \la \partial_{x_i} \|x_i^*-x_0^*\|_p = 0$ into the right-hand side of \eqref{theorem2_3_2-tmp} yields:
\begin{equation}\label{theorem2_3_2}
\begin{split}
&\E\|\nabla F(x_i^k,\xi_i^k)+\la \partial_{x_i} \|x_i^k-x_0^k\|_p\|^2\\
\leq&2\E\|\nabla \E\big[ F(x_i^k,\xi_i^k)\big] - \nabla\E\big[F(x_i^*,\xi_i^k)\big] +\la \partial_{x_i} \|x_i^k-x_0^k\|_p - \la \partial_{x_i} \|x_i^*-x_0^*\|_p\|^2+2\delta_i^2\\
\leq&4\E\|\nabla \E\big[ F(x_i^k,\xi_i^k)\big] - \nabla \E\big[F(x_i^*,\xi_i^k)\big]\|^2 + 4 \la^2\E\| \partial_{x_i} \|x_i^k-x_0^k\|_p - \partial_{x_i} \|x_i^*-x_0^*\|_p\|^2+2\delta_i^2\\
\stackrel{(c)}{\leq}&4\E\|\nabla \E\big[ F(x_i^k,\xi_i^k)\big]-\nabla \E\big[F(x_i^*,\xi_i^k)\big]\|^2+16\la^2d+2\delta_i^2
\end{split}
\end{equation}
where (c) holds true because the $b$-norm of $\partial_{x_i}\|x_i^k-x_0^k\|_p$ (or $\partial_{x_i}\|x_i^*-x_0^*\|_p$) is no larger than $1$ according to Proposition \ref{prop.1}, and thus the absolute value of every element of the $d$-dimensional vector $\partial_{x_i}\|x_i^k-x_0^k\|_p$ (or $\partial_{x_i}\|x_i^*-x_0^*\|_p$) is no larger than $1$.

For the third term at the right-hand side of \eqref{theorem2_3}, since $\E\big[F(x_i,\xi_i^k)\big]$ is strongly convex and has Lipschitz continuous gradients by hypothesis, we have \cite{nesterov2013}:
\begin{equation}\label{theorem2_3_1}
\begin{split}
\E\La\nabla \E\big[F(x_i^k,\xi_i^k)\big]-\nabla \E\big[F(x_i^*,\xi_i^k)\big],x_i^k-x_i^*\Ra\geq\f{\mu_iL_i}{\mu_i+L_i}\E\|x_i^k-x_i^*\|^2+\f{1}{\mu_i+L_i}\E\|\nabla \E\big[F(x_i^k,\xi_i^k)\big]-\nabla \E\big[F(x_i^*,\xi_i^k)\big]\|^2
\end{split}
\end{equation}

Substituting \eqref{theorem2_3_2} and \eqref{theorem2_3_1} into \eqref{theorem2_3}, we have:
\begin{equation}\label{theorem2_4}
\begin{split}
\E\|x_i^{k+1}-x_i^*\|^2\leq& \left(1-\f{2\A^{k+1}\mu_iL_i}{\mu_i+L_i}\right)\E\|x_i^k-x_i^*\|^2+(\A^{k+1})^2(16\la^2d+2\delta_i^2)\\
&-2\A^{k+1}\left(\f{1}{\mu_i+L_i}-2\A^{k+1}\right)\E\|\nabla \E\big[ F(x_i^k,\xi_i^k)\big]-\nabla \E\big[F(x_i^*,\xi_i^k)\big]\|^2\\
&-2\A^{k+1}\E\La\la \partial_{x_i} \|x_i^k-x_0^k\|_p-\la \partial_{x_i} \|x_i^*-x_0^*\|_p,x_i^k-x_i^*\Ra\\
\leq&\left(1- \eta \A^{k+1}\right)\E\|x_i^k-x_i^*\|^2+(\A^{k+1})^2(16\la^2d+2\delta_i^2)\\
&-2\A^{k+1}\E\La\la \partial_{x_i} \|x_i^k-x_0^k\|_p-\la \partial_{x_i} \|x_i^*-x_0^*\|_p,x_i^k-x_i^*\Ra
\end{split}
\end{equation}
where we drop the term of $\E\|\nabla \E\big[ F(x_i^k,\xi_i^k)\big]-\nabla \E\big[F(x_i^*,\xi_i^k)\big]\|^2$ because $\f{1}{\mu_i+L_i}-2\A^{k+1} \geq 0$ according to the step size rule.

\noindent \textbf{Step 2.} From the RSA update \eqref{eq:update2:x:0} at the master, we have:
\begin{equation}\label{theorem2_5}
\begin{split}
 &\E\|x_0^{k+1}-x_0^*\|^2\\
=&\E\bigg\|x_0^k-x_0^*-\A^{k+1} \bigg(\n f_0(x_0^k)+\la\sum_{i\in \R}\partial_{x_0} \|x_0^k-x_i^k\|_p+\la\sum_{j\in\mathcal{B}}\partial_{x_0} \|x_0^k-z_j^k\|_p\bigg)\bigg\|^2\\
=&\E\|x_0^k-x_0^*\|^2+(\A^{k+1})^2\E\bigg\|\n f_0(x_0^k)+\la\sum_{i\in \R}\partial_{x_0} \|x_0^k-x_i^k\|_p+\la\sum_{j\in\mathcal{B}}\partial_{x_0} \|x_0^k-z_j^k\|_pv\bigg\|^2\\
&-2\A^{k+1}\E\La\n f_0(x_0^k)+\la\sum_{i\in \R}\partial_{x_0} \|x_0^k-x_i^k\|_p,x_0^k-x_0^*\Ra -2\A^{k+1}\E\La\la\sum_{j\in\mathcal{B}} \partial_{x_0} \|x_0^k-z_j^k\|_p,x_0^k-x_0^*\Ra.
\end{split}
\end{equation}

For the second term at the right-hand side of \eqref{theorem2_5}, we have:
\begin{equation}\label{theorem2_5_2-tmp}
\begin{split}
&\E\bigg\|\n f_0(x_0^k)+\la\sum_{i\in \R}\partial_{x_0} \|x_0^k-x_i^k\|_p+\la\sum_{j\in\mathcal{B}}\partial_{x_0} \|x_0^k-z_j^{k}\|_p\bigg\|^2\\
\leq&2\E\bigg\|\n f_0(x_0^k)+\la\sum_{i\in \R}\partial_{x_0} \|x_0^k-x_i^k\|_p\bigg\|^2+2\la^2\E\bigg\|\sum_{j\in\mathcal{B}}\partial_{x_0} \|x_0^k-z_j^{k}\|_p\bigg\|^2.
\end{split}
\end{equation}
Since every element of the $d$-dimensional vector $\partial_{x_0} \|x_0^k-z_j^{k}\|_p$ is within $[-1,1]$, it holds:
\begin{equation}\label{theorem2_5_2-tmp-1}
\begin{split}
\E\bigg\|\sum_{j\in\mathcal{B}}\partial_{x_0} \|x_0^k-z_j^{k}\|_p\bigg\|^2 \leq q^2d.
\end{split}
\end{equation}
For $\E\|\n f_0(x_0^k)+\la\sum_{i\in \R}\partial_{x_0} \|x_0^k-x_i^k\|_p\|^2$, we insert the optimality condition of \eqref{eq:tv-p} with respect to $x_0$, namely, $\nabla f_0(x_0^*) + \la \sum_{i \in \mathcal{R}} \partial_{x_0} \|x_i^*-x_0^*\|_p = 0$ to obtain:
\begin{equation}\label{theorem2_5_2-tmp-2}
\begin{split}
&\E\bigg\|\n f_0(x_0^k)+\la\sum_{i\in \R}\partial_{x_0} \|x_0^k-x_i^k\|_p\bigg\|^2 \\
\leq&\E\bigg\|\n f_0(x_0^k) - \nabla f_0(x_0^*) +\la\sum_{i\in \R}\partial_{x_0} \|x_0^k-x_i^k\|_p - \la \sum_{i \in \mathcal{R}} \partial_{x_0} \|x_i^*-x_0^*\|_p\bigg\|^2 \\
\leq&2\E\|\n f_0(x_0^k)-\n f_0(x_0^*)\|^2+2\la^2\E\bigg\|\sum_{i\in \R}\partial_{x_0} \|x_0^k-x_i^k\|_p-\sum_{i\in \R}\partial_{x_0} \|x_0^*-x_i^*\|_p\bigg\|^2 \\
\leq&2\E\|\n f_0(x_0^k)-\n f_0(x_0^*)\|^2+8\la^2r^2d
\end{split}
\end{equation}
Substituting \eqref{theorem2_5_2-tmp-1} and \eqref{theorem2_5_2-tmp-2} into \eqref{theorem2_5_2-tmp} yields:
\begin{equation}\label{theorem2_5_2}
\begin{split}
&\E\|\n f_0(x_0^k)+\la\sum_{i\in \R}\partial_{x_0} \|x_0^k-x_i^k\|_p+\la\sum_{j\in\mathcal{B}}\partial_{x_0} \|x_0^k-z_j^{k}\|_p\|^2\\
\leq&4\E\|\n f_0(x_0^k)-\n f_0(x_0^*)\|^2+16\la^2r^2d+2\la^2q^2d
\end{split}
\end{equation}

For the third term at the right-hand side of \eqref{theorem2_5}, we insert $\nabla f_0(x_0^*) + \la \sum_{i \in \mathcal{R}} \partial_{x_0} \|x_i^*-x_0^*\|_p = 0$, the optimality condition of \eqref{eq:tv-p} with respect to $x_0$, and obtain:
\begin{equation}\label{theorem2_5_1}
\begin{split}
&\E\La\n f_0(x_0^k)+\la\sum_{i\in \R}\partial_{x_0} \|x_0^k-x_i^k\|_p,x_0^k-x_0^*\Ra\\
=   &\E\La\la\sum_{i\in \R}\partial_{x_0} \|x_0^k-x_i^k\|_p-\la\sum_{i\in \R}\partial_{x_0} \|x_0^*-x_i^*\|_p,x_0^k-x_0^*\Ra +\E\La\n f_0(x_0^k)-\n f_0(x_0^*),x_0^k-x_0^*\Ra\\
\stackrel{(d)}{\geq}&\E\La\la\sum_{i\in \R}\partial_{x_0} \|x_0^k-x_i^k\|_p-\la\sum_{i\in \R}\partial_{x_0} \|x_0^*-x_i^*\|_p,x_0^k-x_0^*\Ra +\f{\mu_0L_0}{\mu_0+L_0}\E\|x_0^k-x_0^*\|^2+\f{1}{\mu_0+L_0}\E\|\n f_0(x_0^k)-\n f_0(x_0^*)\|^2
\end{split}
\end{equation}
where (d) is due to the fact that $f_0$ is strongly convex and has Lipschitz continuous gradients (cf. Assumptions \ref{ass:conv} and \ref{ass:lip}).

For the last term at the right-hand side of \eqref{theorem2_5}, it holds for any $\epsilon > 0$ that:
\begin{equation}\label{theorem2_5_3}
\begin{split}
2\E\Big\La\la\sum_{j\in\mathcal{B}}\partial_{x_0} \|x_0^k-z_j^k\|_p,x_0^k-x_0^*\Big\Ra\leq&\epsilon\E\|x_0^k-x_0^*\|^2+\f{\la^2}{\epsilon}\E\Big\|\sum_{j\in\mathcal{B}}\partial_{x_0} \|x_0^k-z_j^k\|_p\Big\|^2\\
\leq&\epsilon\E\|x_0^k-x_0^*\|^2+\f{\la^2q^2d}{\epsilon}.
\end{split}
\end{equation}

Substituting \eqref{theorem2_5_2}, \eqref{theorem2_5_1} and \eqref{theorem2_5_3} into \eqref{theorem2_5}, we have:
\begin{equation}\label{theorem2_6}
\begin{split}
\E\|x_0^{k+1}-x_0^*\|^2\leq&\left(1-(\f{2\mu_0L_0}{\mu_0+L_0}-\epsilon)\A^{k+1}\right)\E\|x_0^{k-1}-x_0^*\|^2+(\A^{k+1})^2(16\la^2r^2d+2\la^2q^2d) +\f{\A^{k+1}\la^2q^2d}{\epsilon}\\
&-\A^{k+1}\left(\f{2}{\mu_0+L_0}-4\A^{k+1}\right)\E\|\n f_0(x_0^k)-\n f_0(x_0^*)\|^2\\
&-2\A^{k+1}\E\La\la\sum_{i\in \R}\partial_{x_0} \|x_0^k-x_i^k\|_p-\la\sum_{i\in \R}\partial_{x_0} \|x_0^*-x_i^*\|_p,x_0^k-x_0^*\Ra\\
\leq&(1-\eta\A^{k+1})\E\|x_0^k-x_0^*\|^2+(\A^{k+1})^2(16\la^2r^2d+2\la^2q^2d)+\f{\A^{k+1}\la^2q^2d}{\epsilon}\\
&-2\A^{k+1}\E\La\la\sum_{i\in \R}\partial_{x_0} \|x_0^k-x_i^k\|_p-\la\sum_{i\in \R}\partial_{x_0} \|x_0^*-x_i^*\|_p,x_0^k-x_0^*\Ra.
\end{split}
\end{equation}
We drop the term of $\E\|\n f_0(x_0^k)-\n f_0(x_0^*)\|^2$ because $\f{1}{\mu_0+L_0}-2\A^{k+1} \geq 0$ according to the step size rule.

\noindent \textbf{Step 3.} Denote $g_p(x)=\sum_{i\in \R}\|x_i-x_0\|_p$. Since $g_p(x)$ is convex, we have:
\begin{equation}\label{theorem2_7}
\begin{split}
  & \La \partial_x g_p(x^k)- \partial_x g_p(x^*),x^k-x^*\Ra \\
= & \sum_{i\in \R} \La \partial_{x_i} \|x_i^k-x_0^k\|_p- \partial_{x_i} \|x_i^*-x_0^*\|_p,x_i^k-x_i^*\Ra +  \sum_{i\in \R} \La\partial_{x_0} \|x_0^k-x_i^k\|_p- \partial_{x_0} \|x_0^*-x_i^*\|_p,x_0^k-x_0^*\Ra \geq 0.
\end{split}
\end{equation}

Summing up \eqref{theorem2_4} for all $i\in\R$, adding \eqref{theorem2_6} and substituting \eqref{theorem2_7}, we have:
\begin{equation}\label{theorem2_8}
\begin{split}
\E\|x^{k+1}-x^*\|^2\leq&(1-\eta\A^{k+1})\E\|x^k-x^*\|^2+(\A^{k+1})^2(16\la^2rd+16\la^2r^2d+2\la^2q^2d+2\sum_{i\in\R}\delta_i^2)+\f{\A^{k+1}\la^2q^2d}{\epsilon}\\
= &(1-\eta\A^{k+1})\E\|x^k-x^*\|^2+(\A^{k+1})^2 \Delta_0 +\A^{k+1} \Delta_2
\end{split}
\end{equation}
where for simplicity we denote:
\begin{equation}\label{app-eq.const}
\Delta_0 = 16\la^2rd+16\la^2r^2d+2\la^2q^2d+2\sum_{i\in\R}\delta_i^2 \quad \text{and} \quad \Delta_2 = \f{\la^2q^2d}{\epsilon}.
\end{equation}

%According to the step size rule $\A^{k+1} = \frac{\overline{\alpha}}{k+k_0}$, where $\overline{\alpha} > \frac{1}{\eta}$ and $k_0 \geq \frac{\overline{\alpha}}{\tau}$ with
%$$\tau = \min\left\{ \min_{i \in \mathcal{R}} \frac{1}{2(\mu_i + L_i)}, \frac{1}{2(\mu_0 + L_0)}, \frac{1}{\eta} \right\},$$
%we rewrite \eqref{theorem2_8} as:
%
%\begin{equation}\label{theorem2_9}
%\E\|x^{k+1}-x^*\|^2\leq \left(1-\f{\eta\overline{\alpha}}{k+k_0} \right)\E\|x^k-x^*\|^2+\f{\overline{\alpha}^2 \Delta_0}{(k+k_0)^2}+\f{\Delta_2elta_2}{k+k_0}.
%\end{equation}
%
%Note that $1-\f{\eta\overline{\alpha}}{k+k_0} \geq 0$ for any $k \geq 0$.

According to the step size rule $\A^{k+1} = \min\{\underline{\alpha},\frac{\overline{\alpha}}{k+1}\}$, there exists a smallest integer $k_0$ satisfying $\underline{\alpha} \geq \frac{\overline{\alpha}}{k_0+1}$ such that $\A^{k+1} = \underline{\alpha}$ when $k < k_0$ and $\A^{k+1} = \frac{\overline{\alpha}}{k+1}$ when $k \geq k_0$. Then for all $k < k_0$, \eqref{theorem2_8} becomes:
\begin{equation}\label{theorem2_new_1}
\E\|x^{k+1}-x^*\|^2\leq (1-\eta \underline{\alpha})\E\|x^k-x^*\|^2+(\underline{\alpha})^2 \Delta_0+\underline{\alpha} \Delta_2, \quad \forall k < k_0.
\end{equation}
By definitions $\eta = \min\{ \min_{i \in \mathcal{R}} \frac{2\mu_i L_i}{\mu_i + L_i}, \frac{2\mu_0 L_0}{\mu_0 + L_0} - \epsilon \}$ and $\underline{\alpha} = \min\{ \min_{i \in \mathcal{R}} \frac{1}{2(\mu_i + L_i)}, \frac{1}{2(\mu_0 + L_0)} \}$, $\eta \underline{\alpha} \in (0, \frac{1}{4})$. Applying telescopic cancellation to \eqref{theorem2_new_1} through time $0$ to $k < k_0$ yields:
\begin{equation}\label{theorem2_new_2}
\E\|x^{k+1}-x^*\|^2 \leq (1-\eta \underline{\alpha})^{k} \E\|x^0-x^*\|^2+ \frac{1}{\eta} (\underline{\alpha} \Delta_0+ \Delta_2), \quad \forall k < k_0
\end{equation}
%
%
%\begin{equation}\label{theorem2_new_2}
%\begin{split}
%\E\|x^{k+1}-x^*\|^2 & \leq (1-\eta \underline{\alpha})^{k} \E\|x^0-x^*\|^2+ \frac{1}{\eta} (\underline{\alpha} \Delta_0+ \Delta_2) \\
%& \leq \E\|x^0-x^*\|^2+ \frac{1}{\eta} (\underline{\alpha} \Delta_0+ \Delta_2), \quad \forall k \leq k_0-1
%\end{split}
%\end{equation}
%

For all $k \geq k_0$, \eqref{theorem2_8} becomes:
\begin{equation}\label{theorem2_new_3}
\E\|x^{k+1}-x^*\|^2\leq (1-\frac{\eta \overline{\alpha}}{k+1})\E\|x^k-x^*\|^2+\frac{\overline{\alpha}^2 \Delta_0}{(k+1)^2}+\frac{\overline{\alpha}\Delta_2}{k+1}, \quad \forall k \geq k_0.
\end{equation}
Note that $1-\frac{\eta \overline{\alpha}}{k+1} \in (\frac{3}{4},1)$ when $k \geq k_0$ because $\frac{\eta \overline{\alpha}}{k+1} \leq \frac{\eta \overline{\alpha}}{k_0+1} \leq \eta \underline{\alpha} < \frac{1}{4}$.
We use induction to prove that:
\begin{equation}\label{theorem2_10}
\E\|x^{k+1}-x^*\|^2\leq \frac{\Delta_1}{k+1}+ \overline{\alpha}\Delta_2, \quad \forall k \geq k_0
\end{equation}
where
$$\Delta_1 = \max\left\{ \frac{\overline{\alpha}^2 \Delta_0}{\eta \overline{\alpha} -1}, (k_0+1) \E\|x^{k_0}-x^*\|^2 + \frac{\overline{\alpha}^2 \Delta_0}{k_0+1} \right\}.$$
When $k=k_0$, \eqref{theorem2_10} holds because by \eqref{theorem2_new_3} it follows:
\begin{equation}\label{theorem2_10-nnn1}
\begin{split}
\E\|x^{k_0+1}-x^*\|^2 & \leq (1-\frac{\eta \overline{\alpha}}{k_0+1})\E\|x^{k_0}-x^*\|^2+\frac{\overline{\alpha}^2 \Delta_0}{(k_0+1)^2}+\frac{\overline{\alpha}\Delta_2}{k_0+1} \\
& \leq \E\|x^{k_0}-x^*\|^2+\frac{\overline{\alpha}^2 \Delta_0}{(k_0+1)^2}+ \overline{\alpha}\Delta_2 \\
& \leq \frac{\Delta_1}{k_0+1} + \overline{\alpha}\Delta_2.
\end{split}
\end{equation}
Then, we assume that \eqref{theorem2_10} holds for a certain $k \geq k_0$ and establish an upper bound for $\E\|x^{k+2}-x^*\|^2$ as:
\begin{equation}\label{theorem2_10-nnn2}
\begin{split}
\E\|x^{k+2}-x^*\|^2\leq&(1-\f{\eta\overline{\alpha}}{k+2})\E\|x^{k+1}-x^*\|^2+\f{\overline{\alpha}^2 \Delta_0}{(k+2)^2}+\f{\overline{\alpha}\Delta_2}{k+2}\\
\leq&(1-\f{\eta\overline{\alpha}}{k+2}) \frac{\Delta_1}{k+1}+\f{\overline{\alpha}^2\Delta_0}{(k+2)^2}+\overline{\alpha}\D + (1-\eta\overline{\alpha}) \frac{\Delta_2}{k+2}\\
\stackrel{(e)}{\leq} &(1-\f{\eta\overline{\alpha}}{k+2}) \frac{\Delta_1}{k+1}+\f{\overline{\alpha}^2\Delta_0}{(k+2)^2}+\overline{\alpha}\D\\
\stackrel{(f)}{\leq}&(1-\f{\eta\overline{\alpha}}{k+2}) \frac{\Delta_1}{k+1}+\f{(\eta \overline{\alpha}-1)\Delta_1}{(k+2)^2}+\overline{\alpha}\D\\
\leq&(1-\f{\eta\overline{\alpha}}{k+2}) \frac{\Delta_1}{k+1}+\f{(\eta \overline{\alpha}-1)\Delta_1}{(k+1)(k+2)}+\overline{\alpha}\D\\
\leq&\f{1}{k+2}\Delta_1+ \overline{\alpha}\D
\end{split}
\end{equation}
where (e) uses the fact that $\overline{\alpha} > \frac{1}{\eta}$, and (f) follows from $\Delta_1 \geq \frac{\overline{\alpha}^2 \Delta_0}{\eta \overline{\alpha} -1}$. This completes the induction as well as the proof.

\section{Convergence of RSA with ${\cal O}(1/\sqrt{k})$ Step Size}
Define the objective function in \eqref{eq:tv-p} as:
\begin{equation}\label{app-eq:tv-p}
h_p(x):=\sum_{i\in\mathcal{R}}\Big(\mathbb{E}[F(x_i,\xi_i)]+\lambda||x_i-x_0||_p\Big)+ f_0(x_0).
\end{equation}
We have the following theorem for RSA with ${\cal O}(1/\sqrt{k})$ step size.

\vspace{0.5em}

\begin{theorem}\label{app-theorem2}
	Suppose that Assumptions \ref{ass:conv}, \ref{ass:lip} and \ref{ass:bound} hold. Set the step size of $\ell_p$-norm RSA ($p\geq 1$) as $\A^{k+1} = \min\{\underline{\alpha},\frac{\overline{\alpha}}{\sqrt{k+1}}\}$, where $\underline{\alpha} = \min \{ \min_{i \in \mathcal{R}} \frac{\mu_i}{4L_i^2}, \frac{\mu_0-\epsilon}{4L_0^2}\}$, $\overline{\alpha} > 0$, and $\epsilon$ is any constant within $(0, \mu_0)$, then we have:
	\begin{equation}\label{app-theorem2_new_2-main}
	\E\bigg[h_p(\bar{x}^k)-h_p(x^*)\bigg]\leq\frac{\E\|x^0-x^*\|^2+\Delta_0\sum_{\tau=0}^k(\A^{\tau+1})^2}{2\sum_{\tau=0}^k\A^{\tau+1}}
	+\frac{\Delta_2}{2}={\cal O}\Big(\frac{\log k}{\sqrt{k}}\Big)+\frac{\Delta_2}{2}
	\end{equation}
	where $\bar{x}^k$ is the running average solution
	$$\bar{x}^k = \frac{\sum_{\tau=0}^k \alpha^{\tau+1}x^\tau}{\sum_{\tau=0}^k \alpha^{\tau+1}},$$
	while $\Delta_0$ and $\D ={\cal O}(\la^2q^2)$ are constants defined as
	$$\Delta_0 = 16\la^2rd+16\la^2r^2d+2\la^2q^2d+2\sum_{i\in\R}\delta_i^2 \quad \text{and} \quad \Delta_2 = \f{\la^2q^2d}{\epsilon}.$$
\end{theorem}

\vspace{0.5em}

\noindent \textit{Proof.} For those equalities and inequalities that also appear in the proof of Theorem \ref{theorem2}, we shall directly cite them. The proof contains the following steps.

%From the RSA update \eqref{eq:update2:x:i} at every regular worker $i$, we have:
%%
%\begin{equation}\label{theorem2_1}
%\begin{split}
%\E\|x_i^{k+1}-x_i^*\|^2=&\E\big\|x_i^k-x_i^*-\A^{k+1}\big(\nabla F(x_i^k,\xi_i^k)+\la\partial\|x_i^k-x_0^k\|_p\big)\big\|^2\\
%=&\E\|x_i^k-x_i^*\|^2+(\A^{k+1})^2\E\|\nabla F(x_i^k,\xi_i^k)+\la \partial\|x_i^k-x_0^k\|_p\|^2\\
%&-2\A^{k+1}\E\La\nabla F(x_i^k,\xi_i^k),x_i^k-x_i^*\Ra-2\A^{k+1}\E\La\la \partial\|x_i^k-x_0^k\|_p,x_i^k-x_i^*\Ra
%\end{split}
%\end{equation}
%%
%Since $x_i^k$ is independent with $\xi_i^k$, it follows that:
%%
%\begin{equation}\label{theorem2_2}
%\E\big[\La\nabla F(x_i^k,\xi_i^k),x_i^k-x_i^*\Ra\big] = \E\big[ \La \nabla \E\big[F(x_i^k,\xi_i^k)\big], x_i^k-x_i^* \Ra\big].
%\end{equation}

\noindent \textbf{Step 1.} From the RSA update \eqref{eq:update2:x:i} at every regular worker $i$, corresponding to \eqref{theorem2_3}, we have:
\begin{equation}\label{app-theorem2_3}
\begin{split}
\E\|x_i^{k+1}-x_i^*\|^2=&\E\|x_i^k-x_i^*\|^2+(\A^{k+1})^2\E\|\nabla F(x_i^k,\xi_i^k)+\la \partial_{x_i} \|x_i^k-x_0^k\|_p\|^2\\
&-2\A^{k+1}\E\La\nabla \E\big[F(x_i^k,\xi_i^k)\big],x_i^k-x_i^*\Ra-2\A^{k+1}\E\La\la \partial_{x_i} \|x_i^k-x_0^k\|_p,x_i^k-x_i^*\Ra\\
\stackrel{(a)}{\leq}&\E\|x_i^k-x_i^*\|^2+(\A^{k+1})^2\E\|\nabla F(x_i^k,\xi_i^k)+\la \partial_{x_i} \|x_i^k-x_0^k\|_p\|^2\\
&-2\A^{k+1}\E(\E\big[F(x_i^k,\xi_i^k)\big]-\E\big[F(x_i^*,\xi_i^k)\big]+\frac{\mu_i}{2}\|x_i^k-x_i^*\|^2)\\
&-2\A^{k+1}\E\La\la \partial_{x_i} \|x_i^k-x_0^k\|_p,x_i^k-x_i^*\Ra
\end{split}
\end{equation}
where (a) is due to the strong convexity of $\E\big[ F(x_i,\xi_i^k)\big]$.

For the second term at the right-hand side of \eqref{app-theorem2_3}, we have:
\begin{equation}\label{app-theorem2_3_2-tmp}
\begin{split}
    &\E\|\nabla F(x_i^k,\xi_i^k)+\la \partial_{x_i} \|x_i^k-x_0^k\|_p\|^2 \\
=   &\E\|\nabla \E\big[ F(x_i^k,\xi_i^k)\big]+\la \partial\|x_i^k-x_0^k\|_p+\n F(x_i^k,\xi_i^k)-\n\E\big[ F(x_i^k,\xi_i^k)\big]\|^2\\
\leq&2\E\|\nabla \E\big[ F(x_i^k,\xi_i^k)\big]+\la \partial_{x_i} \|x_i^k-x_0^k\|_p\|^2+2\E[\|\n F(x_i^k,\xi_i^k)-\n\E\big[ F(x_i^k,\xi_i^k)\big]\|^2]\\
\stackrel{(b)}{\leq}&2\E\|\nabla \E\big[ F(x_i^k,\xi_i^k)\big]+\la \partial_{x_i} \|x_i^k-x_0^k\|_p\|^2+2\delta_i^2
\end{split}
\end{equation}
where (b) is due to the bounded variance given by Assumption \ref{ass:bound}. Plugging the optimality condition $\E\big[\nabla F(x_i^*,\xi_i^k)\big] + \la \partial_{x_i} \|x_i^*-x_0^*\|_p = 0$ into $\E\|\nabla \E\big[ F(x_i^k,\xi_i^k)\big]+\la \partial_{x_i} \|x_i^k-x_0^k\|_p\|^2$ in the right-hand side of \eqref{app-theorem2_3_2-tmp} yields:
\begin{equation}\label{app-theorem2_3_2}
\begin{split}
&\E\|\nabla F(x_i^k,\xi_i^k)+\la \partial\|x_i^k-x_0^k\|_p\|^2\\
\leq&2\E\|\nabla \E\big[ F(x_i^k,\xi_i^k)\big] - \E\big[F(x_i^*,\xi_i^k)\big] +\la \partial\|x_i^k-x_0^k\|_p - \la \partial\|x_i^*-x_0^*\|_p\|^2+2\delta_i^2\\
\leq&4\E\|\nabla \E\big[ F(x_i^k,\xi_i^k)\big] - \nabla \E\big[F(x_i^*,\xi_i^k)\big]\|^2 + 4 \la^2\E\| \partial\|x_i^k-x_0^k\|_p - \partial\|x_i^*-x_0^*\|_p\|^2+2\delta_i^2\\
\stackrel{(c)}{\leq}&4\E\|\nabla \E\big[ F(x_i^k,\xi_i^k)\big]-\nabla \E\big[F(x_i^*,\xi_i^k)\big]\|^2+16\la^2d+2\delta_i^2\\
\stackrel{(d)}{\leq}&4L^2_i\E\|x_i^k-x_i^*\|^2+16\la^2d+2\delta_i^2
\end{split}
\end{equation}
where (c) holds true because the $b$-norm of $\partial_{x_i} \|x_i^k-x_0^k\|_p$ (or $\partial_{x_i} \|x_i^*-x_0^*\|_p$) is no larger than $1$ according to Proposition \ref{prop.1}, and thus the absolute value of every element of the $d$-dimensional vector $\partial_{x_i} \|x_i^k-x_0^k\|_p$ (or $\partial_{x_i} \|x_i^*-x_0^*\|_p$) is no larger than $1$, and (d) is due to the Lipschitz continuous gradients of $F(x_i,\xi_i^k)$.
Substituting \eqref{app-theorem2_3_2} into \eqref{app-theorem2_3}, we have:
\begin{equation}\label{app-theorem2_4}
\begin{split}
\E\|x_i^{k+1}-x_i^*\|^2
\leq&\E\|x_i^k-x_i^*\|^2-2\A^{k+1}\E(\E\big[F(x_i^k,\xi_i^k)\big]-\E\big[F(x_i^*,\xi_i^k)\big])\\
&-\A^{k+1}(\mu_i-4\A^{k+1}L_i^2)\E\|x_i^k-x_i^*\|^2+(\A^{k+1})^2(16\la^2d+2\delta_i^2)\\
&-2\A^{k+1}\E\La\la \partial_{x_i} \|x_i^k-x_0^k\|_p,x_i^k-x_i^*\Ra\\
\leq&\E\|x_i^k-x_i^*\|^2-2\A^{k+1}(\E\big[F(x_i^k,\xi_i^k)\big]-\E\big[F(x_i^*,\xi_i^k)\big])\\
&-2\A^{k+1}\E\La\la \partial_{x_i} \|x_i^k-x_0^k\|_p,x_i^k-x_i^*\Ra+(\A^{k+1})^2(16\la^2d+2\delta_i^2)\\
\end{split}
\end{equation}
We drop the term of $\E\|x_i^k-x_i^*\|^2$ because $\mu_i-4\A^{k+1}L_i^2 \geq 0$ according to the step size rule.

\noindent \textbf{Step 2.} From the RSA update \eqref{eq:update2:x:0} at the master, we have:
\begin{equation}\label{app-theorem2_5}
\begin{split}
 &\E\|x_0^{k+1}-x_0^*\|^2\\
=&\E\bigg\|x_0^k-x_0^*-\A^{k+1} \bigg(\n f_0(x_0^k)+\la\sum_{i\in \R}\partial_{x_0} \|x_0^k-x_i^k\|_p+\la\sum_{j\in\mathcal{B}}\partial_{x_0} \|x_0^k-z_j^k\|_p\bigg)\bigg\|^2\\
=&\E\|x_0^k-x_0^*\|^2+(\A^{k+1})^2\E\|\n f_0(x_0^k)+\la\sum_{i\in \R}\partial_{x_0} \|x_0^k-x_i^k\|_p+\la\sum_{j\in\mathcal{B}}\partial_{x_0} \|x_0^k-z_j^k\|_p\|^2\\
&-2\A^{k+1}\E\Big\La\n f_0(x_0^k)+\la\sum_{i\in \R}\partial_{x_0} \|x_0^k-x_i^k\|_p,x_0^k-x_0^*\Big\Ra -2\A^{k+1}\E\La\la\sum_{j\in\mathcal{B}} \partial_{x_0} \|x_0^k-z_j^k\|_p,x_0^k-x_0^*\Ra\\
\end{split}
\end{equation}
which is the same as \eqref{theorem2_5}.

For the second term at the right-hand side of \eqref{app-theorem2_5}, corresponding \eqref{app-theorem2_5_2-tmp}, we have:
\begin{equation}\label{app-theorem2_5_2-tmp}
\begin{split}
&\E\|\n f_0(x_0^k)+\la\sum_{i\in \R}\partial_{x_0} \|x_0^k-x_i^k\|_p+\la\sum_{j\in\mathcal{B}}\partial_{x_0} \|x_0^k-z_j^{k}\|_p\|^2\\
\leq&2\E\|\n f_0(x_0^k)+\la\sum_{i\in \R}\partial_{x_0} \|x_0^k-x_i^k\|_p\|^2+2\la^2\E\|\sum_{j\in\mathcal{B}}\partial_{x_0} \|x_0^k-z_j^{k}\|_p\|^2
\end{split}
\end{equation}
For $\E\|\n f_0(x_0^k)+\la\sum_{i\in \R}\partial_{x_0} \|x_0^k-x_i^k\|_p\|^2$, we insert the optimality condition of \eqref{eq:tv-p} with respect to $x_0$, namely, $\nabla f_0(x_0^*) + \la \sum_{i \in \mathcal{R}} \partial_{x_0} \|x_i^*-x_0^*\|_p = 0$ to obtain:
\begin{equation}\label{app-theorem2_5_2-tmp-2}
\begin{split}
    &\E\|\n f_0(x_0^k)+\la\sum_{i\in \R}\partial_{x_0}\|x_0^k-x_i^k\|_p\|^2 \\
\leq&\E\|\n f_0(x_0^k) - \nabla f_0(x_0^*) +\la\sum_{i\in \R}\partial_{x_0}\|x_0^k-x_i^k\|_p - \la \sum_{i \in \mathcal{R}} \partial_{x_0}\|x_i^*-x_0^*\|_p\|^2 \\
\leq&2\E\|\n f_0(x_0^k)-\n f_0(x_0^*)\|^2+2\la^2\E\|\sum_{i\in \R}\partial_{x_0}\|x_0^k-x_i^k\|_p-\sum_{i\in \R}\partial_{x_0}\|x_0^*-x_i^*\|_p\|^2 \\
\leq&2\E\|\n f_0(x_0^k)-\n f_0(x_0^*)\|^2+8\la^2r^2d\\
\stackrel{(e)}{\leq}&2L_0^2\E\|x_0^k-x_0^*\|^2+8\la^2r^2d
\end{split}
\end{equation}
where (e) is due to the Lipschitz continuous gradients of $f_0(x_0)$. Substituting $\E \|\sum_{j\in\mathcal{B}}\partial_{x_0} \|x_0^k-z_j^{k}\|_p \|^2 \leq q^2d$ in \eqref{theorem2_5_2-tmp-1} and \eqref{app-theorem2_5_2-tmp-2} into \eqref{app-theorem2_5_2-tmp} yields:
\begin{equation}\label{app-theorem2_5_2}
\begin{split}
\E\|\n f_0(x_0^k)+\la\sum_{i\in \R}\partial_{x_0} \|x_0^k-x_i^k\|_p+\la\sum_{j\in\mathcal{B}}\partial_{x_0} \|x_0^k-z_j^{k}\|_p\|^2\leq&4L_0^2\E\|x_0^k-x_0^*\|^2+16\la^2r^2d+2\la^2q^2d
\end{split}
\end{equation}

For the third term at the right-hand side of \eqref{app-theorem2_5}, since $f_0(x_0)$ is strongly convex with constant
$\mu_0$, we have:
\begin{equation}\label{app-theorem2_5_1}
\begin{split}
&\E\La\n f_0(x_0^k)+\la\sum_{i\in \R}\partial_{x_0} \|x_0^k-x_i^k\|_p,x_0^k-x_0^*\Ra\\
\geq&\E (f_0(x_0^k)- f_0(x_0^*)+\frac{\mu_0}{2}\|x_0^k-x_0^*\|^2)+\E\La\la\sum_{i\in \R}\partial_{x_0} \|x_0^k-x_i^k\|_p,x_0^k-x_0^*\Ra
\end{split}
\end{equation}

For the last term at the right-hand side of \eqref{app-theorem2_5}, it holds for any $\epsilon > 0$ that:
\begin{equation}\label{app-theorem2_5_3}
\begin{split}
2\E\Big\La\la\sum_{j\in\mathcal{B}}\partial_{x_0} \|x_0^k-z_j^k\|_p,x_0^k-x_0^*\Big\Ra\leq&\epsilon\E\|x_0^k-x_0^*\|^2+\f{\la^2}{\epsilon}\E\Big\|\sum_{j\in\mathcal{B}}\partial_{x_0} \|x_0^k-z_j^k\|_p\Big\|^2\\
\leq&\epsilon\E\|x_0^k-x_0^*\|^2+\f{\la^2q^2d}{\epsilon}.
\end{split}
\end{equation}

Substituting \eqref{app-theorem2_5_2}, \eqref{app-theorem2_5_1} and \eqref{app-theorem2_5_3} into \eqref{app-theorem2_5}, we have:
\begin{equation}\label{app-theorem2_6}
\begin{split}
\E\|x_0^{k+1}-x_0^*\|^2\leq&\E\|x_0^{k}-x_0^*\|^2+(\A^{k+1})^2(16\la^2r^2d+2\la^2q^2d)+\f{\A^{k+1}\la^2q^2d}{\epsilon}\\
&-2\A^{k+1}\E [f_0(x_0^k)- f_0(x_0^*)]-2\A^{k+1}\E\La\la\sum_{i\in \R}\partial_{x_0} \|x_0^k-x_i^k\|_p,x_0^k-x_0^*\Ra\\
&-\A^{k+1}(\mu_0-4L_0^2\A^{k+1}-\epsilon)\E\|x_0^k-x_0^*\|^2\\
\leq&\E\|x_0^{k}-x_0^*\|^2+(\A^{k+1})^2(16\la^2r^2d+2\la^2q^2d)+\f{\A^{k+1}\la^2q^2d}{\epsilon}\\
&-2\A^{k+1}\E [f_0(x_0^k)- f_0(x_0^*)]-2\A^{k+1}\E\La\la\sum_{i\in \R}\partial_{x_0} \|x_0^k-x_i^k\|_p,x_0^k-x_0^*\Ra\\
\end{split}
\end{equation}
where we drop the term of $\E\|x_0^k-x_0^*\|^2$ because $\mu_0-4L_0^2\A^{k+1}-\epsilon \geq 0$ according to the step size rule.

\noindent \textbf{Step 3.} Using the convexity of $\|x_i-x_0\|_p$, we have:
\begin{equation}\label{app-theorem2_7}
\sum_{i\in\mathcal{R}}\La \partial_{x_i} \|x_0^k-x_i^k\|_p,x_i^k-x_i^*\Ra+ \sum_{i\in\mathcal{R}} \La \partial_{x_0} \|x_0^k-x_i^k\|_p,x_0^k-x_0^*\Ra\geq \sum_{i\in\mathcal{R}}\|x_i^k-x_0^k\|_p-\sum_{i\in\mathcal{R}}\|x_i^*-x_0^*\|_p.
\end{equation}

Summing up \eqref{app-theorem2_4} for all $i\in\R$, as well as combining \eqref{app-theorem2_6} and \eqref{app-theorem2_7}, we have:
\begin{equation}\label{app-theorem2_8}
\begin{split}
2\A^{k+1}\E(h_p(x^k)-h_p(x^*))\leq\E\|x^k-x^*\|^2-\E\|x^{k+1}-x^*\|^2+(\A^{k+1})^2\Delta_0+\A^{k+1}\Delta_2.
\end{split}
\end{equation}
where $\Delta_0$ and $\Delta_2$ are constants defined in \eqref{app-eq.const}.
Summing up \eqref {app-theorem2_8} for all times $k$, we have:
\begin{equation}\label{app-theorem2_9}
2\sum_{\tau=0}^k \A^{\tau+1}E\bigg[\sum_{\tau=0}^k\frac{\A^{\tau+1}}{\sum_{\tau=0}^k\A^{\tau+1}}(h_p(x^\tau)-h_p(x^*))\bigg]\leq\E\|x^0-x^*\|^2+\Delta_0\sum_{\tau=0}^k(\A^{\tau+1})^2+\Delta_2\sum_{\tau=0}^k\A^{\tau+1}\end{equation}
Since $h_p(x)$ is convex, we have:
\begin{equation}\label{app-theorem2_10}
\sum_{\tau=0}^k\frac{\A^{\tau+1}}{\sum_{\tau=0}^k\A^{\tau+1}}(h_p(x^\tau)-h_p(x^*))\geq h_p(\bar{x}^k)-h_p(x^*).
\end{equation}
Substituting \eqref{app-theorem2_10} to \eqref{app-theorem2_9} yields:
\begin{equation}\label{app-theorem2_11}
\E\bigg[h_p(\bar{x}^k)-h_p(x^*)\bigg]\leq\frac{\E\|x^0-x^*\|^2+\Delta_0\sum_{\tau=0}^k(\A^{\tau+1})^2}{2\sum_{\tau=0}^k\A^{\tau+1}}
+\frac{\Delta_2}{2}.
\end{equation}

\section{Proof of Theorem \ref{theorem3}}

\noindent \textit{Proof.} When $\lambda \geq \lambda_0$, combining Theorem \ref{theorem1} and Theorem \ref{theorem2} directly yields \eqref{eq:the3}. When $0 < \lambda < \lambda_0$, we have: $$\E [\|x^k - [\tilde{x}^*]\|^2 ] \leq 2\E [\|x^k - x^*\|^2] + 2\E [\|x^* - [\tilde{x}^*]\|^2]$$
where the inequality follows from $(a+b)^2 \leq 2a^2 + 2b^2$. By Theorem \ref{theorem2} and $\E [\|x^* - [\tilde{x}^*]\|^2] \leq \Delta_3$, \eqref{eq:the3-1} holds true.

\end{document}